\documentclass[sigconf]{acmart}
\usepackage{xurl}
\usepackage{xcolor}

\definecolor{cl1-1}{HTML}{2271B2} 
\definecolor{cl1-2}{HTML}{F748A5} 
\definecolor{cl1-3}{HTML}{359B73} 

\definecolor{cl2-0}{HTML}{003D30} 
\definecolor{cl2-1}{HTML}{790149} 
\definecolor{cl2-2}{HTML}{00306F} 
\definecolor{cl2-3}{HTML}{65019F} 
\definecolor{cl2-4}{HTML}{005FCC} 
\definecolor{cl2-5}{HTML}{5A000F} 
\definecolor{cl2-6}{HTML}{A40122} 
\definecolor{cl2-7}{HTML}{F60239} 
\definecolor{cl2-8}{HTML}{FF6E3A} 

\AtBeginDocument{%
  \providecommand\BibTeX{{%
    \normalfont B\kern-0.5em{\scshape i\kern-0.25em b}\kern-0.8em\TeX}}}

\copyrightyear{2021}
\acmYear{2021}
\setcopyright{acmlicensed}\acmConference[GeoSearch'21]{1st ACM SIGSPATIAL International Workshop on Searching and Mining Large Collections of Geospatial Data }{November 2, 2021}{Beijing, China}
\acmBooktitle{1st ACM SIGSPATIAL International Workshop on Searching and Mining Large Collections of Geospatial Data (GeoSearch'21), November 2, 2021, Beijing, China}
\acmPrice{15.00}
\acmDOI{10.1145/3486640.3491392}
\acmISBN{978-1-4503-9123-8/21/11}
\begin{document}

\title{gtfs2vec - Learning GTFS Embeddings for comparing Public Transport Offer in Microregions}

\author{Piotr Gramacki}
\email{p.gramacki@gmail.com}
\orcid{0000-0002-4587-5586}
\affiliation{%
  \institution{Department of Computational Intelligence, Wrocław University of Science and Technology}
  \city{Wrocław}
  \country{Poland}}
  
\author{Szymon Woźniak}
\email{swozniak6@gmail.com}
\orcid{0000-0002-2047-1649}
\affiliation{%
  \institution{Department of Computational Intelligence, Wrocław University of Science and Technology}
  \city{Wrocław}
  \country{Poland}}

\author{Piotr Szymański}
\email{piotr.szymanski@pwr.edu.pl}
\orcid{0000-0002-7733-3239}
\affiliation{%
  \institution{Department of Computational Intelligence, Wrocław University of Science and Technology}
  \city{Wrocław}
  \country{Poland}}


\begin{abstract}
We selected 48 European cities and gathered their public transport timetables in the GTFS format. We utilized Uber's H3 spatial index to divide each city into hexagonal micro-regions. Based on the timetables data we created certain features describing the quantity and variety of public transport availability in each region. Next, we trained an auto-associative deep neural network to embed each of the regions. Having such prepared representations, we then used a hierarchical clustering approach to identify similar regions. To do so, we utilized an agglomerative clustering algorithm with a euclidean distance between regions and Ward's method to minimize in-cluster variance. Finally, we analyzed the obtained clusters at different levels to identify some number of clusters that qualitatively describe public transport availability. We showed that our typology matches the characteristics of analyzed cities and allows succesful searching for areas with similar public transport schedule characteristics.
\end{abstract}



\begin{teaserfigure}
\centering
  \includegraphics[width=0.24\textwidth]{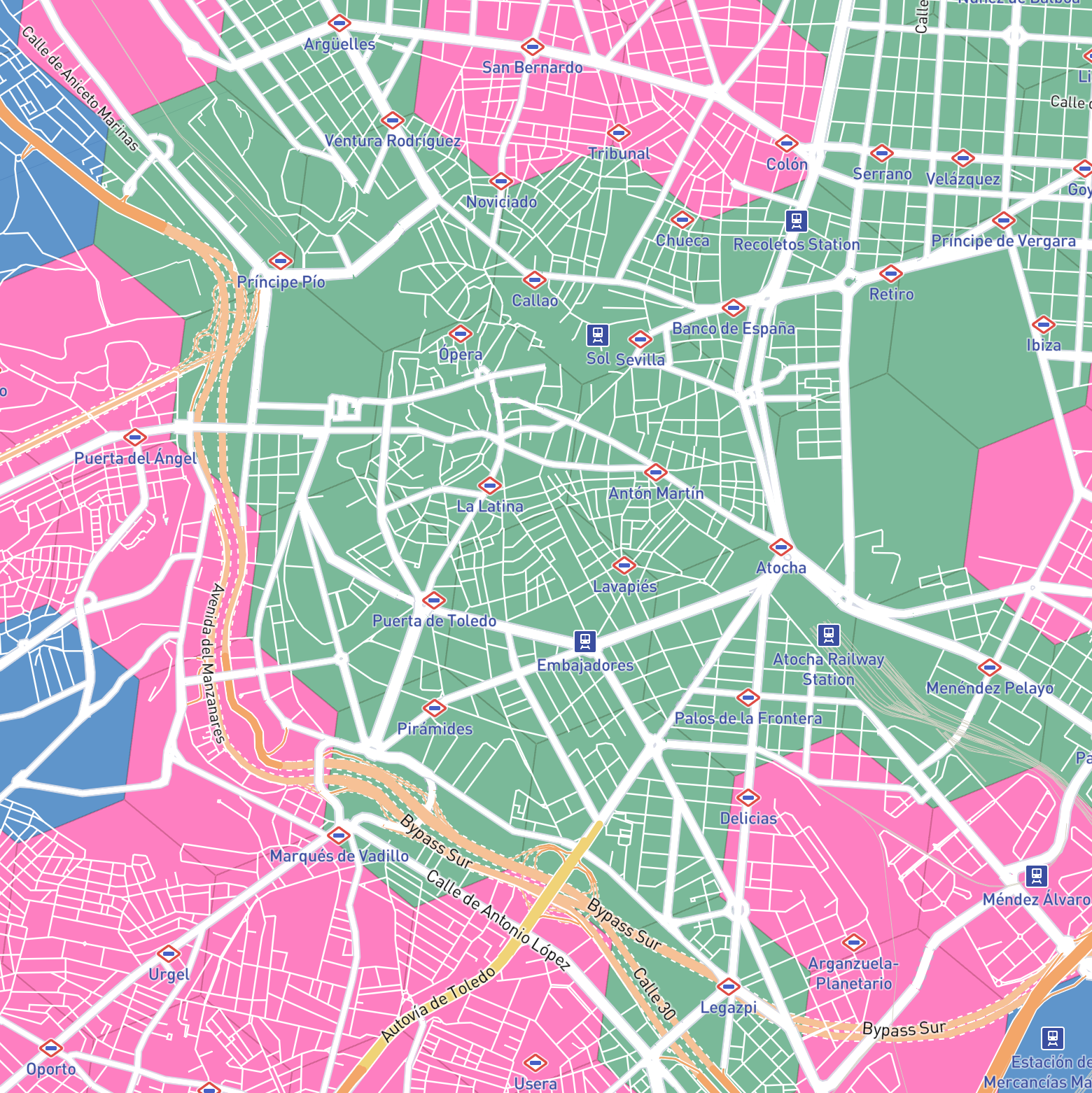}
  \hfill
  \includegraphics[width=0.24\textwidth]{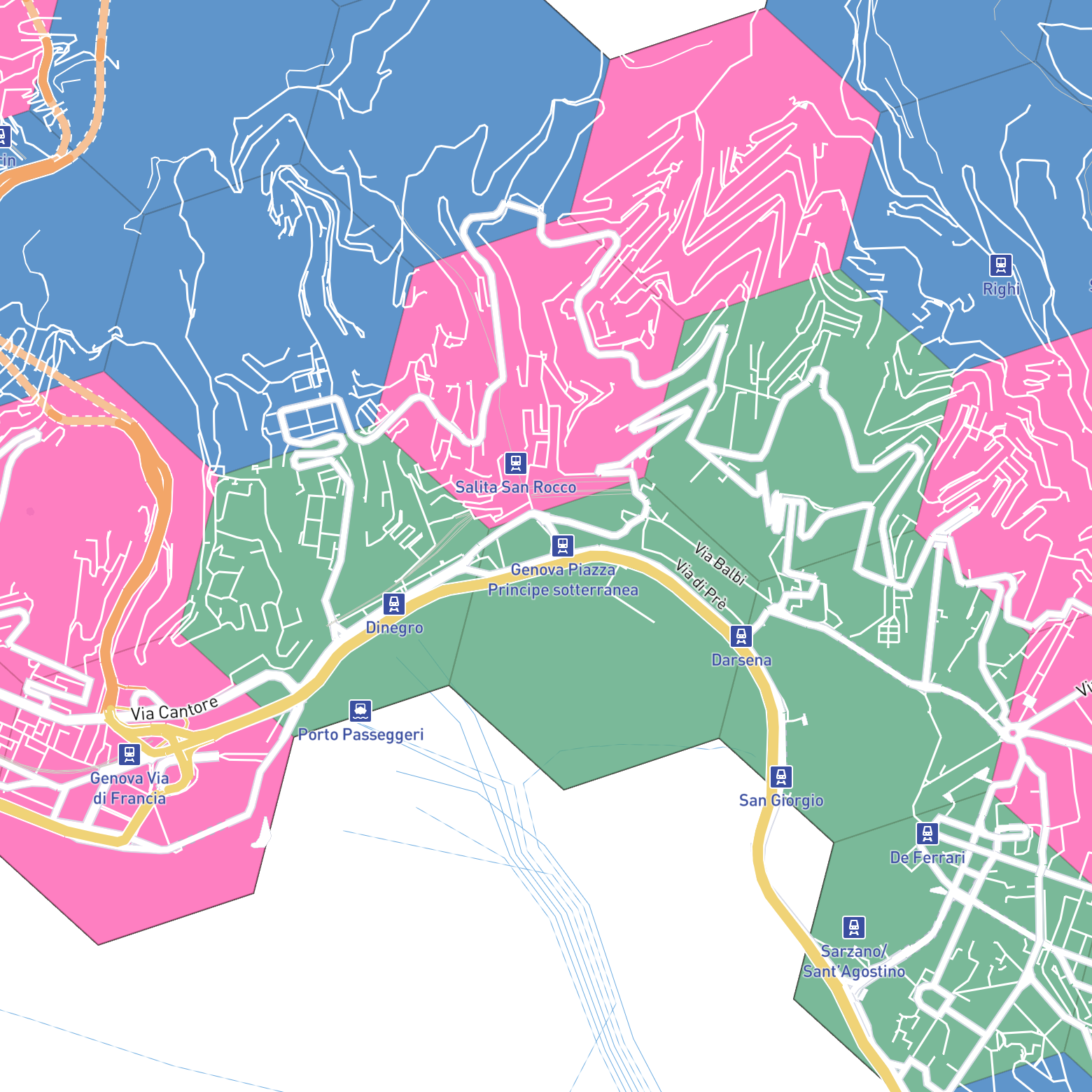}
  \hfill
  \includegraphics[width=0.24\textwidth]{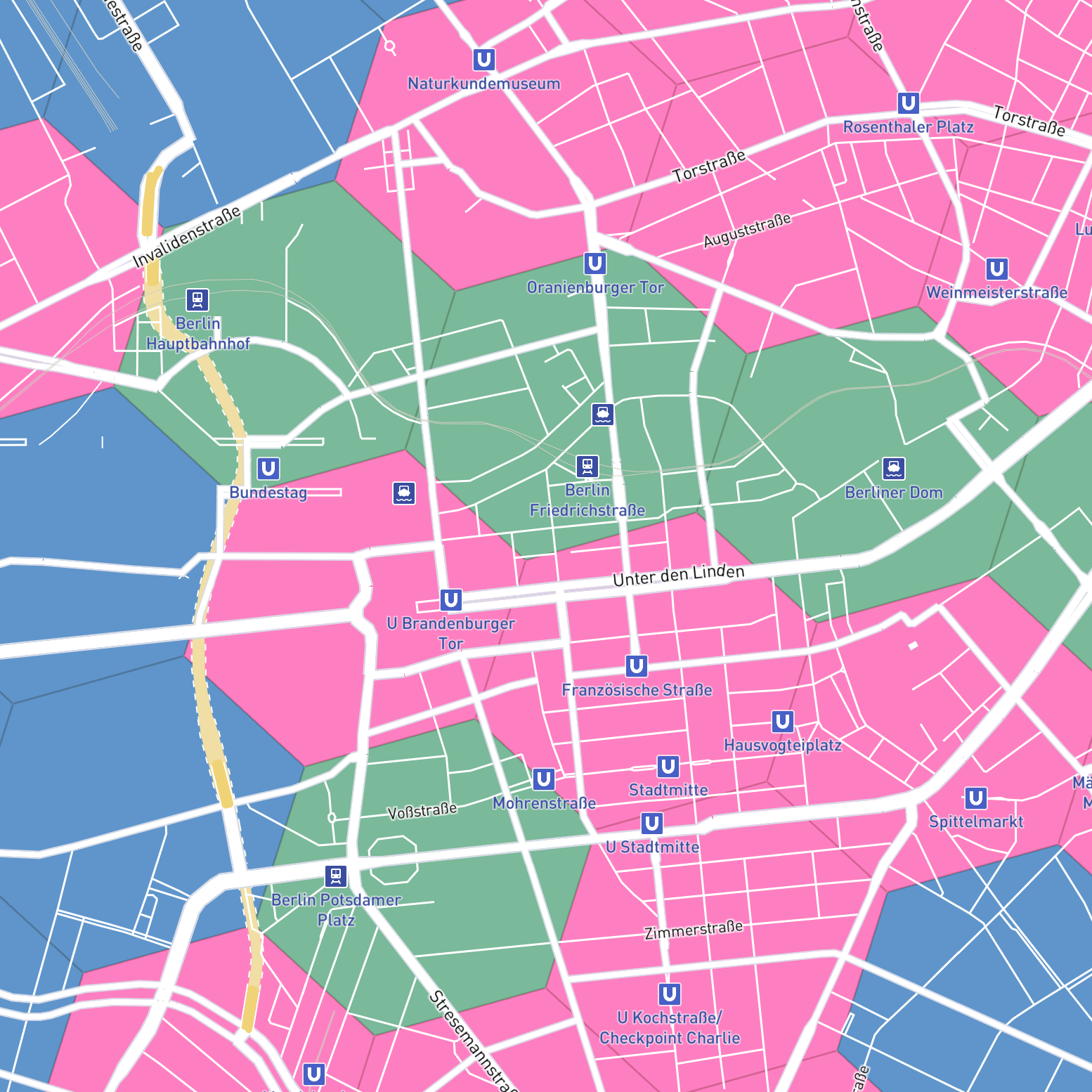}
  \hfill
  \includegraphics[width=0.24\textwidth]{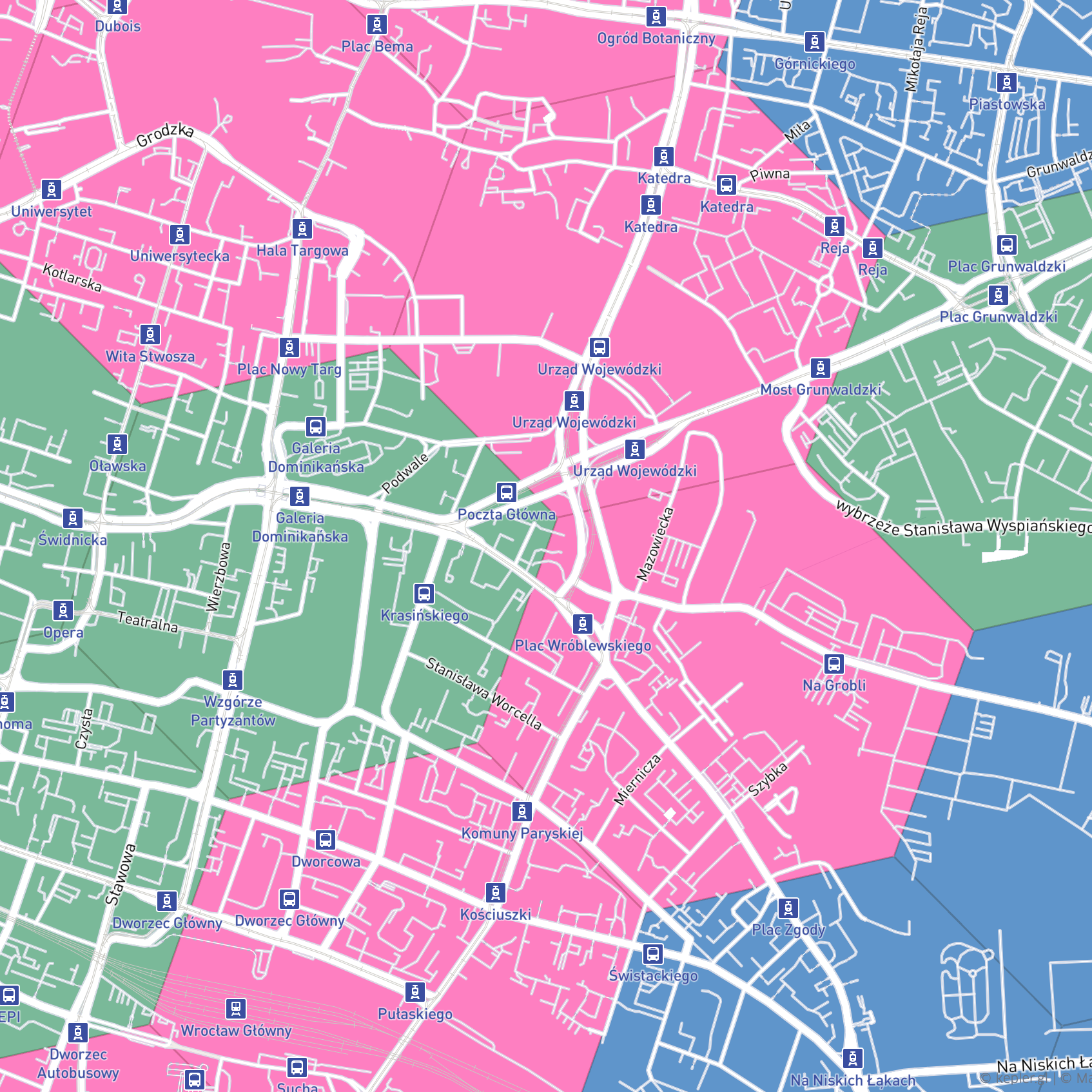}
  \caption{Examples of regions with different types of public transport offer}
\end{teaserfigure}

\maketitle

\section{Introduction}
\label{chapter:intorduction}

Spatial data are available widely for many cities around the world. They are used in solving urban-related tasks such as traffic flow prediction, house price estimation, road network optimization, identification of accident locations, managing transportation issues, and many others. They may also be a valuable source of information in comparing so-called micro-regions in the city. For this, however, it is necessary to have a method of representation of those data. In this paper, we propose the first GTFS representation learning approach which allows comparing regions in terms of the quality of public transport offer. To confirm that the proposed representations are useful, we perform in depth exploratory analysis of cluster obtained based on regions' vector similarity. Using the obtained clusters, we formulate a public transport offer typology and follow with examples of how the obtained vectors can be used to search a database of micro-regions for regions with similar public transport typology.

\subsection{Representation learning in spatial data science}

Term representation learning is used to describe a set of techniques for the automated discovery of features from raw data. One of the methods from this area is embedding. It can be defined as a task of finding an optimal function $f: D \rightarrow \rm{I\!R}^{d}$, which transforms data $D$ into $d$-dimensional vector representation. The main advantage which is gained when translating pure spatial data to latent space is the possibility to define a measure of distance between objects. This makes comparing micro-regions in the city in a certain spatial context possible. Finally, one can perform exploration in such a new space, which may reveal previously unnoticed patterns or features present in the city. Representation learning approach has been widely applied to spatial data in recent years. Some of the works focus on proposing an embedding method which is then tested on a variety of tasks \cite{zone2vec,venue2vec,hrnr}, while other attempt to solve given tasks and for this purpose, they apply representation learning techniques \cite{loc2Vec,cape,ze-mob}. They show an improvement when compared to other methods used for solving such tasks \cite{cape,zone2vec,ze-mob,venue2vec,hrnr}. Moreover, the obtained embedding spaces are often shown to be interpretable \cite{loc2Vec,venue2vec}. Embedding methods for spatial data examined in this review draw inspiration from other areas where the representation learning approach is widely used. The majority of them is based on natural language processing and word2vec \cite{word2vec} method \cite{cape,zone2vec,ze-mob}. Others used existing methods for graph embedding \cite{venue2vec,hrnr} or methods from the domain of image processing \cite{loc2Vec}. This variety shows that there is a visible and strong trend for applying representation learning for spatial data.

Public transport is picked up in the literature mostly in tasks related to general delays estimation. Most recent methods apply embedding techniques to improve the quality of delays estimating in public transport \cite{Shoman2020,Barnes2020}. Delays were also used as a factor when comparing regions around a city \cite{Raghothama2016}. The task of analyzing and comparing regions in a city in terms of public transport availability is not covered widely in the literature. Methods from the domain of network science were successfully applied to this task \cite{VonFerber2008}, however, this approach cannot be easily transferred to micro-regions comparison. Utilization of travel times \cite{FayyazS.2017} brings valuable insight into the subject of public transport availability and can be transferred to use micro-regions in place of stops. However, the computation costs of this approach are significant, which opens an opportunity for proposing another method that could be easier to apply to a large number of cities. 

\subsection{Public transport schedules}

One of the urban types of data which to the best of the authors' knowledge was not deeply explored in the context of various machine learning or data mining tasks is public transport schedules. Even though they are widely available in a unified format for many cities around the world\footnote{\url{https://transitfeeds.com/}}. It is also obvious that public transport availability is an important aspect when comparing regions in the city or comparing cities with each other. Hence public transport quality is often one of deciding factors when choosing a district to live in. Analysis of transport schedules in the sense of their real suitability for city dwellers is not a trivial task. Therefore, providing a method to learn meaningful vector representation of public transport availability for a given micro-region seems to be important for the development of systems with the aforementioned functionality. Ability to analyze an entire city to find the areas which need an improvement in terms of public transportation would be certainly beneficial for urban planners and city authorities. This may also act as a way to compare current public transport infrastructure between cities. This possibility may be used to compare with cities considered as a "gold standard" in the area of public transport organization. Unfortunately, there is not much research conducted to analyze an existing public transport infrastructure in terms of a static analysis of public transport schedules. Many of them focus mainly on planning optimal routes e.g. between different destinations. If they analyze existing networks they often focus on the prediction of delays or calculation of the estimated time of arrival (ETA). However, exploring similarities between regions of the city or similarities between different cities is still a relatively unexplored area.

General Transit Feed Specification (GTFS) is a unified format for static public transport schedules introduced by Google \cite{gtfs}. It was first used to incorporate transit data to Google Maps and over the years became de facto a standard format for transit feed data. It contains information about stops, routes, and times or arrival for each stop on a route. In addition to this basic set of information, the GTFS format allows specifying the precise route which the vehicle takes. It can also store additional information about a station such as levels, pathways, or fare information. 

\subsection{Identification of micro-regions}

A method of splitting a given area into micro-regions is a crucial aspect of our work. There are many ways of achieving this goal. This section will go through the most popular methods and will justify the choice made in this paper.


The simplest solution is to divide a given area into regular shapes based on a map projection. It is simple and intuitive but may result in some inconveniences. Firstly, such division may depend on map scale, and therefore may be inconsistent between different cities. Secondly, it heavily relies on map projection which is inaccurate because the Earth is a sphere that cannot be precisely mapped to a flat surface. Lastly, such division is not easily repeatable as it may depend on the selected area's borders.

Manual pointing of micro-region is the most accurate solution. It allows to incorporate domain knowledge and creates a division suitable for a given task. It makes it possible to divide a city based on different characteristics e.g. on population density and/or incorporate administrative borders within the city.
An obvious disadvantage of such an approach is that it is labor-intensive and specific to one single area. Therefore it is not suitable for applications that will incorporate different regions of different cities, like the one presented in this paper. 

A spatial index makes it possible to divide a given area into grid cells, identifiable by a unique index. Many spatial indexes work hierarchically, meaning that each cell is composed of smaller cells. Examples of such systems are Google's S2 \cite{s2} based on rectangles and Uber's H3 \cite{h3} which uses hexagons. In this work, the H3 library will be used to partition cities into micro-regions. The reasons behind this decision are presented in the following section.

\subsubsection{H3 hierarchical spatial index}

Uber's solution uses hexagons which is an effective way of space division. One of its advantage over using rectangles or triangles is that all neighbors are in the same distance. Another advantage is the hierarchical character of this system. This allows selecting a resolution that is optimal for micro-region identification and makes it possible to consider sub-regions for each micro-region. 
Hierarchy in H3 is implemented as 16 levels of resolution. On level 0 there are 122 base cells covering an entire planet. The smallest hexagons on level 15 have a diameter of length 1m. 

\begin{figure}[t]
    \centering
    \includegraphics[width=0.45\textwidth]{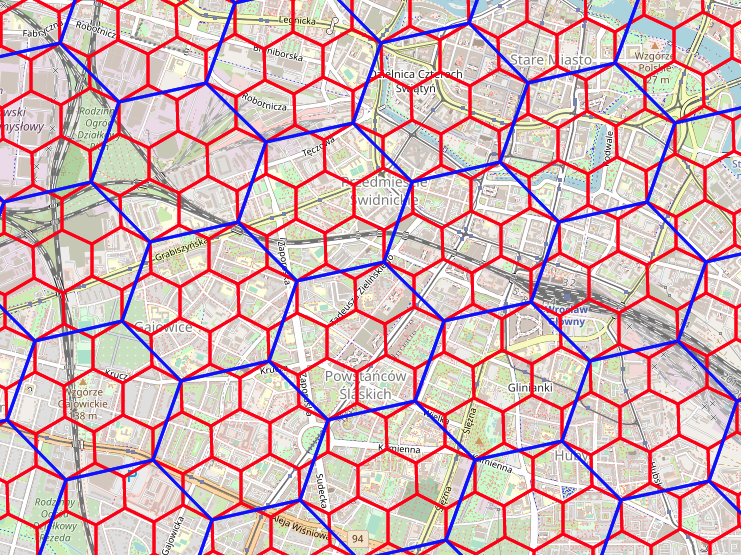}
    \caption{H3 cells - comparison of resolution 8 and 9}
    \Description{Image presents a comparison between resolution 8 and 9 in Uber's H3 spatial index illustrated on a part of Wrocław. Each of bigger hexagon roughly contains 7 smaller.}
    \label{fig:wroclaw-8-9}
\end{figure}

For the task of regions embedding concerning public transport, two seem to be a good fit - 8 and 9. They are respectively about 900m and 350m in diameter. The bigger one covers an area roughly by main roads, whereas resolution 9 is close to a single quarter of buildings. The difference is presented for a small part of Wrocław in Figure \ref{fig:wroclaw-8-9}.

We analyzed how are stops distributed in regions on both resolutions. Using resolution 8 means that there are regions with a lot of stops in them (over 30), whereas resolution 9 results in the majority of regions with no stop in them. Based on that, we selected a resolution 8 further work.

The remainder of this paper is organized as follows: Section 2 presents a proposed solution for features engineering and embedding model, Section 3 reports dataset utilized in this work and results of typology identification, Section 4 summarizes whole work and presents future works. We release the experimental code for this paper alongside obtained embeddings in a dedicated Github repository\footnote{\url{https://github.com/pwr-inf/gtfs2vec}}.

\section{Proposed solution}

This section provides a detailed view of the proposed solution to the problem of public transport accessibility representation and comparison between cities. It covers a feature engineering process and normalization strategy. After that, the architecture and learning process of an embedding model for micro-regions representations learning is discussed in detail. 

\subsection{Features engineering}

As presented in the first section, the GTFS format consists of multiple files and is organized around a list of stop times for routes and stops. Such format is impossible to be used directly in most ML/DL models. Therefore, we proposed feature engineering work to prepare data in the correct format. This section describes this process and justifies the applied solutions. 

From a perspective of a single micro-region, two main public transport accessibility metrics can be defined. Quantity and variety of public transport available. Both were a part of the proposed solution.

\subsubsection{Quantity related features}

To describe the quantity of public transport available for a given micro-region, we calculated the sum of trips in a given period. To perform those calculations, we used the \textit{gtfs-kit} library \cite{gtfs-kit}. It provides methods for the calculation of trips counts for each stop at a given time-frequency. Those numbers were later summed up across an entire micro-region. 

Sometimes multiple stops, which form the same route, are aggregated into a single micro-region. As a result, the same trip will be counted multiple times. However, this is an intentional assumption, because those features are not meant to have a meaning of counting vehicles leaving a micro-region. Instead, they describe in how many ways one can leave an area. The intuition behind this is that having more options to get on a bus/tram inside a region is better than just a single stop in the middle of an area.

\subsubsection{Variety related features}

Another feature which can describe public transport availability is the variety of directions in which one can leave a region. It can be considered on different levels of precision starting from world directions and ending on all regions accessible via public transport with a given number of transfers. For this thesis, an in-between solution was selected. It is described below.

The part of a route description in GTFS format is a \textit{trip\_headsign} column. In practice, this is the destination of a route displayed on a vehicle. They are in most cases either transportation hubs or line ends located most often on the outer ring of the city. This is in a balance between the generality of world directions and the specificity of each region connected via public transport. Hence, to create a feature for a region representation, unique head-signs were counted in a given period, resulting in the total number of unique directions available from a region.

\subsubsection{Time resolution for features and filtering}

Both of the features mentioned above can be calculated for different time windows resulting in a different number of features. For this work, one hour window was selected resulting in 24 quantity and 24 variety features. This resolution seems to be a reasonable choice because it presents changes throughout the day and allows to interpret periods for example as rush hours or nighttime. 

After an analysis of timetables for night lines, we decided to remove them from features of a region. Nighttime transportation availability is significantly different from daytime, as the routes are often longer to minimize the number of vehicles and drivers needed. Moreover, some additional hubs were identified which are not active during the day. Therefore, only 6-22 hours were included as features for each region. As a result, each micro-region was described with a vector $\mathbf{x}$ of length 34.

\subsection{Normalization}

The main purpose of our work was to propose a method of learning representation of public transport accessibility described by features mentioned in the previous section. As an embedding model, we used a neural network, which will be described in the following section. Here we must note that the above-mentioned features which are the main data in our task should be normalized to be able to serve as an input to the neural network.

The values of features describing the number of trips are approximately 10 times larger. This may result in a neural network focusing mostly on those features. Another observation is that values vary for different cities. This was also taken into consideration when proposing a method of normalization.

To achieve the best performance of the embedding model, features will be scaled to the $[0, 1]$ range using \textit{min-max normalization} technique, which is expressed with the formula

\begin{equation}
    \mathbf{x}' = \frac{\mathbf{x} - \min(\mathbf{x})}{\max(\mathbf{x}) - \min(\mathbf{x})}.
\end{equation}

As values of each type (quantity and variety) should maintain relative values during the whole day, normalization will be performed twice: once for all columns with a sum of trips and once for all columns with destinations counts.

Normalization was conducted for all cities combined. This will ensure that differences in public transport available between cities are maintained. This approach should allow comparing the quality of public transport between cities.

\subsection{Embedding method}

Autoencoder is a special type of neural network capable of learning lower- or higher-dimensional embeddings of data. It is trained in an unsupervised manner, which is crucial for this thesis. 


Autoencoder network consists of two networks - encoder and decoder. The encoder network can be treated as a function $\phi: \mathcal{X} \rightarrow \mathcal{Z}$, and decoder as $\psi: \mathcal{Z} \rightarrow \mathcal{X}$. Having input vector $\mathbf{x} \in \mathcal{X}$, encoder network produces an embedding $\phi(\mathbf{x}) = \mathbf{z}$, which is then fed into decoder network to obtain reconstructed input $\psi(\mathbf{z}) = \mathbf{x'}$. Both networks are trained together using backpropagation algorithm, with the loss function for a batch of inputs $\mathbf{X}$ defined as:

\begin{equation}
    \mathcal{L} = \frac{1}{N} \sum_{n=1}^{N} (\mathbf{X}_n - \mathbf{X'}_n)^2,
\end{equation}

which is a \textit{mean squared error loss}. When the training is complete, only the encoder is used for inference.

Historically first applications of autoencoders focused on compressing an input and discovering nonlinear correlations in data can be found in the work \cite{Kramer1991}. It is possible to use nonlinear activation functions, and smaller dimensionality of space $\mathcal{Z}$. In this work, we used the opposite approach. We defined a sparser, 64-dimensional representation with an architecture presented below.

\begin{itemize}
    \item Encoder:
    
    \begin{itemize}
        \item $34 \rightarrow 48$ neurons fully connected layer,
        \item ReLU activation function,
        \item $48 \rightarrow 64$ neurons fully connected layer;
    \end{itemize}
    
    \item Decoder:
    
    \begin{itemize}
        \item $64 \rightarrow 48$ neurons fully connected layer,
        \item ReLU activation function,
        \item $48 \rightarrow 34$ neurons fully connected layer;
    \end{itemize}
    
\end{itemize}

This particular architecture was selected somewhat arbitrarily because in a representation learning domain the goal is to extract representations from raw data and only at the stage of exploratory analysis it is possible to determine whether they are good or not. However, several intuitions are backing up this choice of architecture.
To begin with, the amount of data, which is available for training is not big. The reason for that is the limited availability of timetables data. A city is divided into approximately 300-500 micro-regions, which means that there is not enough data to train a very deep model. Therefore, we decided not to add any more layers to our model.
Moreover, a sparser representation was used to provide more dimensions to differentiate types of public transport. 
Having considered that, a 64-dimensional embedding layer and one hidden layer seem like a reasonable choice. 

\section{Similarities detection}

A final goal of analysis was to define a typology of public transport offer types across multiple cities. To achieve that, we ran clustering multiple times increasing the number of clusters at each step. At each step, we analyzed obtained clusters to determine which number of clusters forms an interpretable typology. The last step will be to describe an obtained typology.

\subsection{Datasets}

In this research we used \textit{Open Mobility Data} \cite{openmobilitydata} service to gather GTFS files for multiple cities in Europe.\footnote{with an exception of Zurich, which was downloaded from \url{https://data.stadt-zuerich.ch/dataset/vbz_fahrplandaten_gtfs}} To limit the number of cities we defined the following inclusion criteria:
\begin{itemize}
    \item city population of at least 200,000 people,
    \item at least 20 routes of public transport,
    \item feed must be complete, meaning no public transport types should be missing,
    \item GTFS file should not be for an entire country (those include intercity trains, which may affect feature engineering process).
\end{itemize}
While processing feeds we encountered some issues, which resulted in an additional exclusion criterion - a \textit{trip\_headsign} column in the trips table must be correct, since it is used to calculate a number of directions available from a region. As a result, we collected data for 48 cities, which are listed in Table \ref{tab:48-cities}.

\begin{table}[ht]
\centering
\caption{List of cities included in research}
\label{tab:48-cities}
\begin{tabular}{@{}lllll@{}}
\toprule
Aachen  & Antwerp  & Athens          & Barcelona       & Belgrade        \\ 
Berlin  & Bilbao   & Brussels        & Budapest        & Bydgoszcz       \\
Cologne & Dublin   & Erfurt          & Espoo           & Florence        \\
Genua   & Gent     & Hamburg         & Helsinki        & Karlsruhe       \\
Krakow  & Lille    & Lublin          & Madrid          & Mannheim        \\
Milan   & Nantes   & Nice            & Oslo            & Oulu            \\
Palermo & Poznan   & Prague          & Radom           & Szczecin        \\
Tampere & Toulouse & \multicolumn{3}{c}{Tricity (Gdansk, Gdynia, Sopot)} \\
Turin   & Valencia & Venice          & Vienna          & Vilnius         \\
Warsaw  & Wroclaw  & Zagreb          & Zurich          & Leipzig         \\ \bottomrule
\end{tabular}
\end{table}

\subsection{Agglomerative clustering}

Having an embedding model trained, we calculated embeddings for all micro-regions. The next step was to analyze the results and finally try to extract valuable information from comparing the cities. This step will use unsupervised methods like distance calculations in embedding space later used in clustering task. As the task solved in this thesis is purely unsupervised in nature, there is no other way to evaluate results other than analyzing obtained space. 

The clustering method requires a distance metric to be defined in obtained embedding space. As a distance metric, cosine and euclidean metrics were considered, following a common practice from literature. A cosine metric finds similarities in \textit{direction} from the origin, which would discard quantity differences between values and focus on a general trend. This would work great in a solution, which aims to find similarities between regions in terms of public transport usage type because it discards differences in for example number of trips. However, in this work, that those differences are important. Therefore, a euclidean distance is used, since it extracts actual distances between samples. It is calculated as

\begin{equation}
    \label{eq:euclidean-distance}
    d(\mathbf{x}, \mathbf{y}) = || \mathbf{x} - \mathbf{y} ||_2
\end{equation}

The last element of the proposed solution for exploratory analysis of obtained embedding space is the clustering method based on a defined distance metric. One of the main goals of this analysis was to propose a multi-level classification in terms of public transport accessibility and quality. This was a reason for choosing a hierarchical approach to clustering, namely an \textit{agglomerative clustering}. 


This is an iterative method, which starts by placing all samples in separate clusters. Then, at each step, two nearest clusters are merged, until a given number of clusters is obtained or another stop criterion is met. To select clusters to merge, Ward's method \cite{ward} was used, because it minimizes intra-cluster variance. This feature is beneficial when identifying a well-differentiated typology. 


To better understand a character of obtained clusters we used two aggregated features: a) \textit{sum\_trips} which represents the sum of trips from an entire day (within 6-22 limits), b) \textit{directrions\_whole\_day} which represents all directions available during a day (from 6 to 22).
They represent public transport availability in a region from whole day perspective. This is a good measure to justify differences between clusters from a distance. 

On the first level of typology, we identified three major types of areas. They were called \textcolor{cl1-1}{suburban}, \textcolor{cl1-2}{mid-city}, and \textcolor{cl1-3}{hubs} for this work. Clearly, those clusters describe a general availability of public transport, where \textcolor{cl1-1}{suburban} has relatively low availability and \textcolor{cl1-2}{mid-city} has high. \textcolor{cl1-3}{Hubs} are typical outliers with very rich public transport offer.

\begin{figure}[h]
    \centering
    \includegraphics[width=0.43\textwidth]{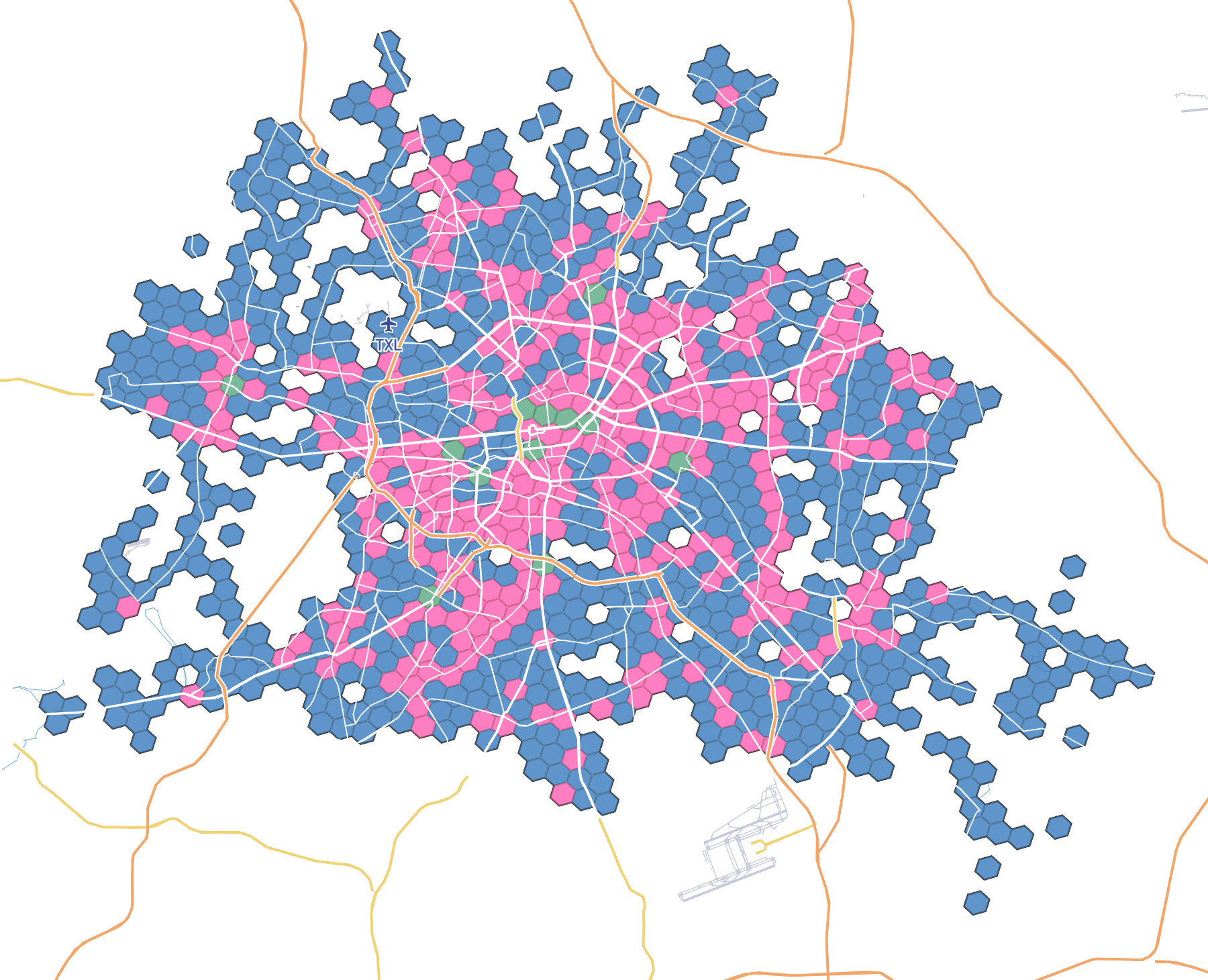}
    \caption{First level typology visualized for Berlin.}
    \label{fig:berlin-clusters1}
\end{figure}

\begin{figure}[h]
    \centering
    \includegraphics[width=0.43\textwidth]{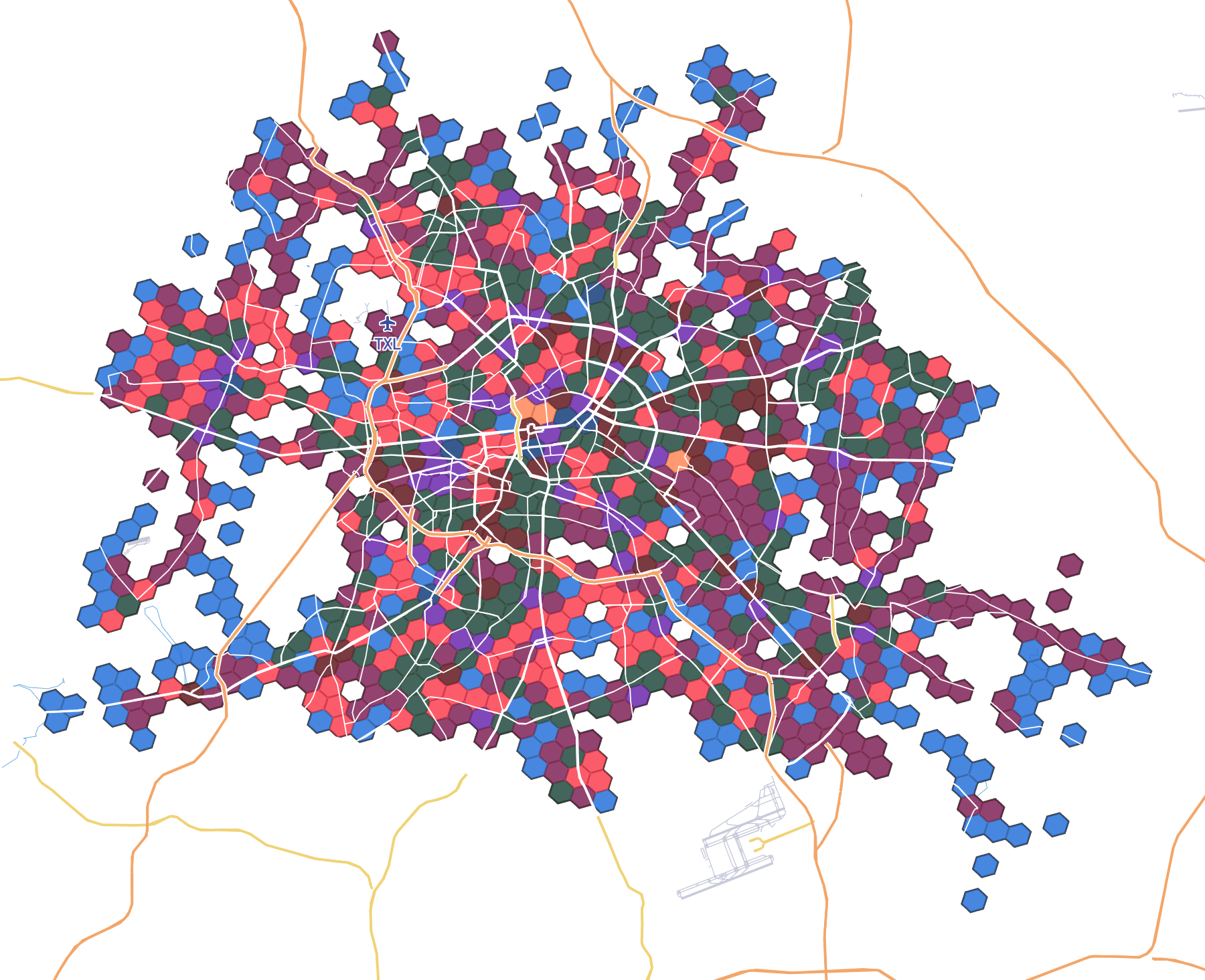}
    \caption{Second level typology visualized for Berlin.}
    \label{fig:berlin-clusters2}
\end{figure}

\begin{figure*}[t]
    \centering
    \includegraphics[width=0.4\textwidth]{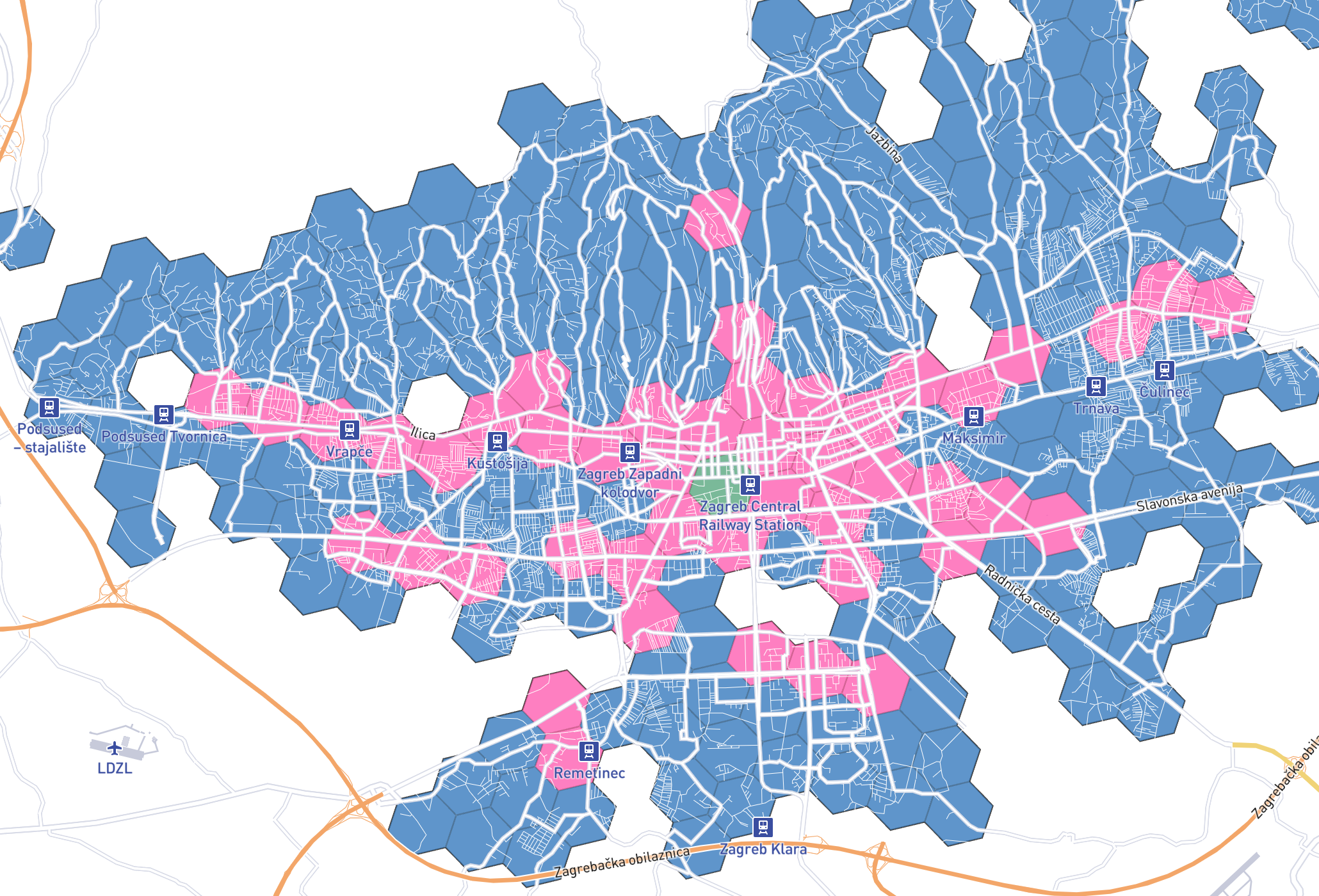}
    \includegraphics[width=0.4\textwidth]{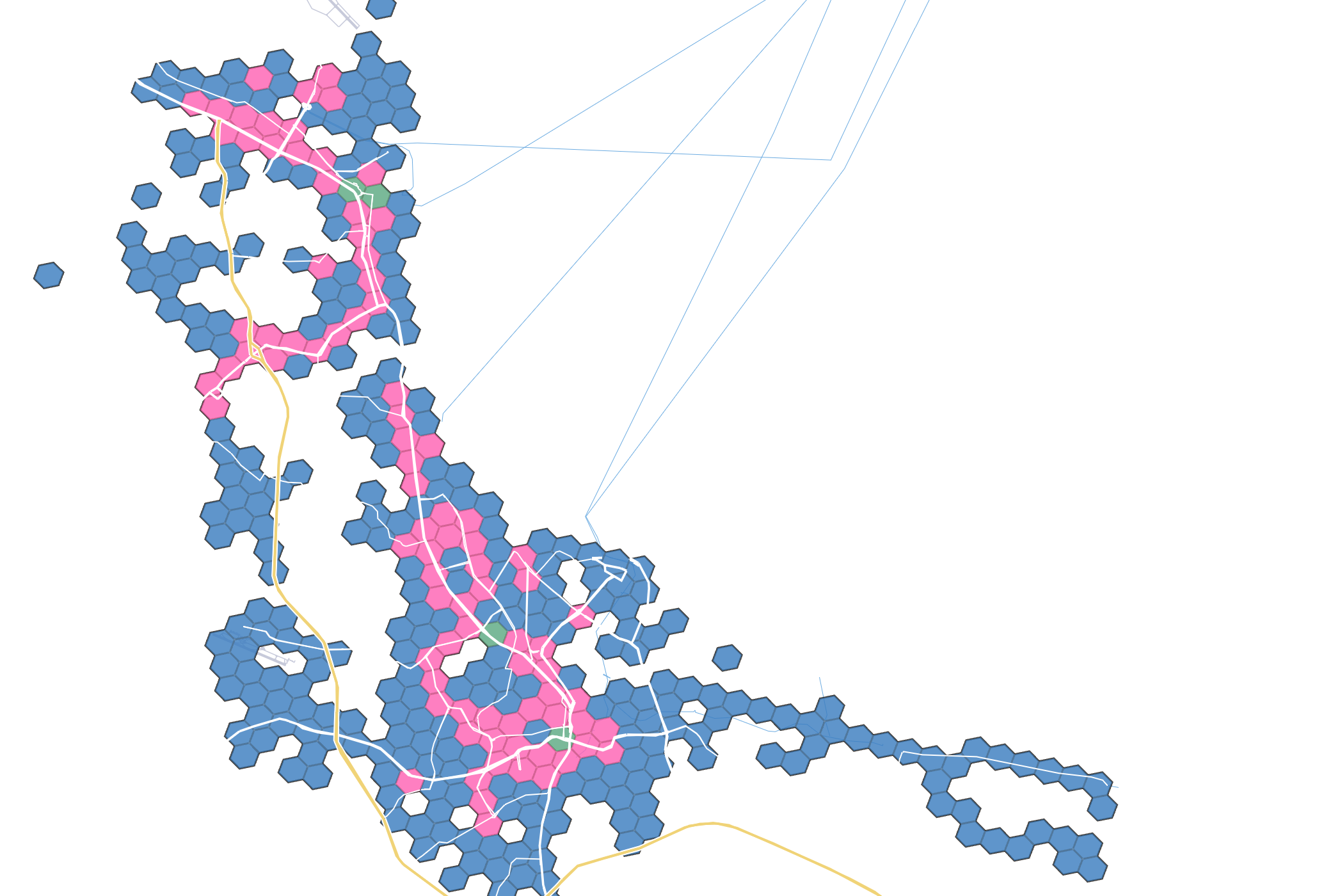}
    \caption{Line cities organized alongside a train line - Zagreb and Tricity (Gdansk, Gdynia, Sopot)}
    \label{fig:line-cities}
\end{figure*}

\subsection{Typology definition}

The second level utilized the division of each type into three more. Each of the level one types was divided differently. From \textcolor{cl1-1}{suburban} areas, we extracted regions with high diversity and high quantity of available public transport (respectively clusters \textcolor{cl2-1}{\textbf{1}} and \textcolor{cl2-7}{\textbf{7}}). They can be treated as local transfer points in an area with poorer public transport availability and well-connected parts of suburbs (for example just one bus line but with a high frequency of trips). The remaining regions from \textcolor{cl1-1}{suburban}-type have significantly worse public transport availability.

The division in \textcolor{cl1-2}{mid-city} type was done similarly. Regions with a high frequency of public transport but with limited variety were separated (\textcolor{cl2-3}{cluster \textbf{3}}). Those are regions with very good availability of public transport. The second type separated regions with a relatively big variety of public transport (\textcolor{cl2-5}{cluster \textbf{5}}). Those can serve as transportation hubs and main transfer points in cities with worse quality of public transport network or as local hubs in districts. The remaining cluster includes regions with city \textcolor{cl1-3}{hubs} which separated differently. One of the clusters shows significantly better public transport availability (\textcolor{cl2-2}{cluster \textbf{2}}). It represents hubs with the highest number of trips and directions. Two remaining clusters maintain only one of them on a high level - either number of trips (\textcolor{cl2-6}{cluster \textbf{6}}) or a number of directions (\textcolor{cl2-8}{cluster \textbf{8}}). The first one would mean regions with a dense metro network. The other contains quite often train stations.
Summarized typology can be defined as:

\begin{itemize}
    \item \textcolor{cl1-1}{suburban} areas
    
    \begin{itemize}
        \item \textcolor{cl2-4}{cluster \textbf{4}} - regions with the least amount of public transport available
        \item \textcolor{cl2-1}{cluster \textbf{1}} - parts of suburbs with better variety of public transport - some sort of entry points for suburbs
        \item \textcolor{cl2-7}{cluster \textbf{7}} - regions in suburbs with more trips but with small variety of public transport
    \end{itemize}
    
    \item \textcolor{cl1-2}{mid-city}
    
    \begin{itemize}
        \item \textcolor{cl2-0}{cluster \textbf{0}} - mid-city regions with regular public transport availability
        \item \textcolor{cl2-3}{cluster \textbf{3}} - mid-city regions with a lot of trips every hour on limited number of directions
        \item \textcolor{cl2-5}{cluster \textbf{5}} - regions with very good public transport availability, in cities with worse public transport network they may serve as hubs, in other just as regional hubs
    \end{itemize}
    
    \item \textcolor{cl1-3}{hubs}
        
    \begin{itemize}
        \item \textcolor{cl2-2}{cluster \textbf{2}} - very high intensity hubs
        \item \textcolor{cl2-6}{cluster \textbf{6}} - regions with a lot of trips, but limited directions, most often regions with metro network in a city center
        \item \textcolor{cl2-8}{cluster \textbf{8}} - railway-type hubs with a lot of diversity and less trips per hour
    \end{itemize}
\end{itemize}

Both levels of the typology were visualized on a map which is available online\footnote{ \url{https://kepler.gl/demo/map?mapUrl=https://dl.dropboxusercontent.com/s/qge8lyeinpmm7ud/keplergl_53g73fa.json}}. The map includes three layers for each of the evaluated cities: 
\begin{itemize}
    \item first level typology - with 3 clusters, as depicted for Berlin in Figure\ref{fig:berlin-clusters1};
    \item second level typology - with 9 clusters, as depicted for Berlin  in Figure\ref{fig:berlin-clusters2};
    \item layer with numeric features from the input table, with number of trips or destinations.
\end{itemize}


\begin{figure}[ht]
    \centering
    \includegraphics[width=0.45\textwidth]{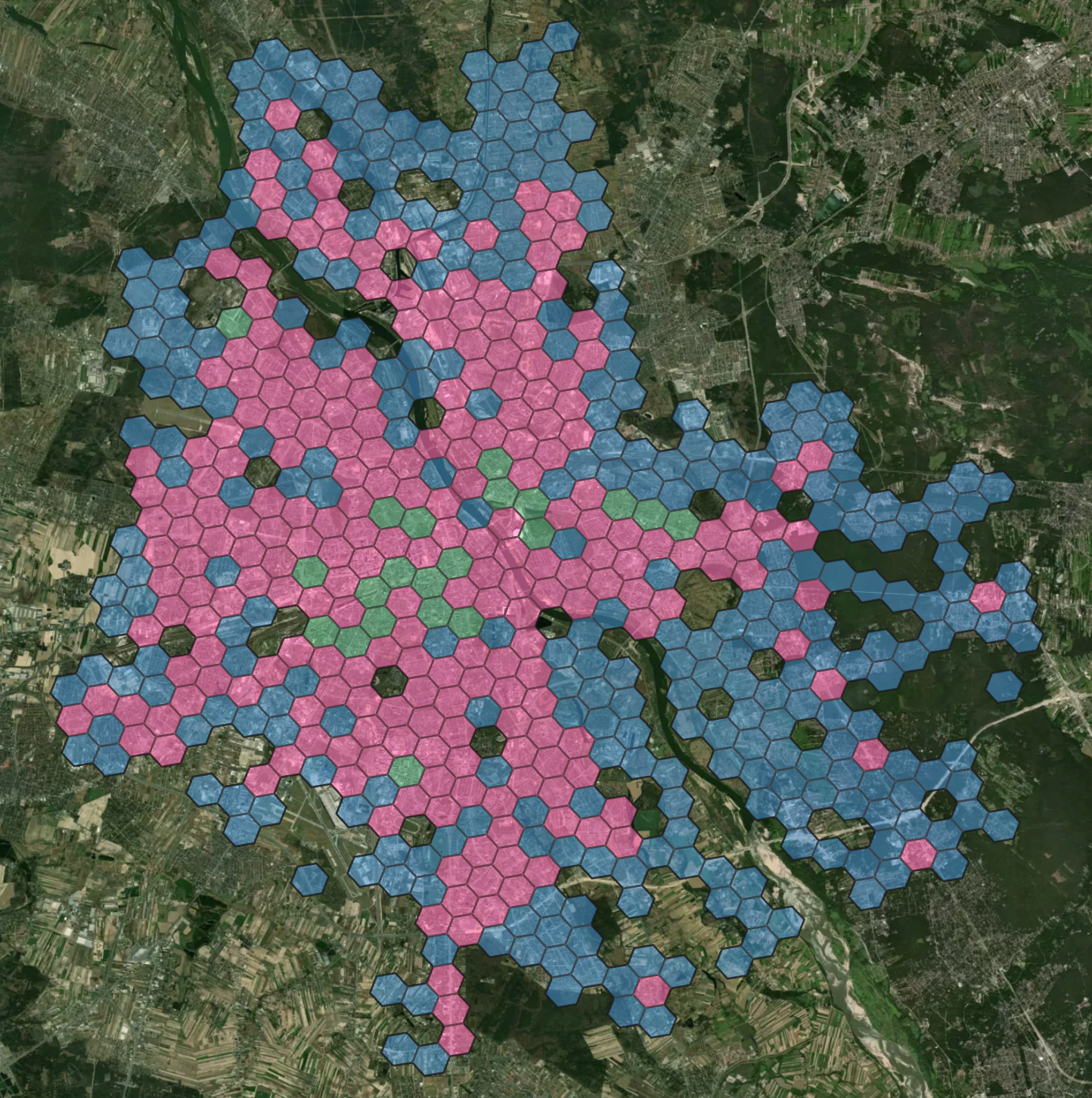}
    \caption{Imbalance in public transport availability between left- and right-bank part of Warsaw}
    \label{fig:imbalance-warsaw}
\end{figure}

\subsection{Similarities between cities}

Our method allows for identification of regions with similar role in different cities. We used both levels of typology to identify those similarities, where the first level allows us to compare cities from a bird's eye view. We used it to compare the types of public transport in \textcolor{cl1-1}{suburban} areas in different cities. First example is a so called "line city" organized around a train line. Two examples of Zagreb and Gdansk are presented in Figure \ref{fig:line-cities}. They both have regions from \textcolor{cl1-2}{mid-city} cluster along a train line, and one or more \textcolor{cl1-3}{hubs} close to main train stations. Parts of a city, which are further from a train line have poorer public transport availability. 

Our vector representations also allow finding imbalances in public transport availability in different parts of a city. Figure \ref{fig:imbalance-warsaw} shows a view of Warsaw, with a visible separation for left- and right-bank parts. Even though, both parts are similar in size, a right-bank part of a city has worse public transport availability. Same observations can be made for Vienna and Budapest (which is visible on maps, which we provide a link to in this publication).

When analyzing a presence of regions assigned to the \textcolor{cl1-3}{hubs} cluster on first level of typology, we identified three types of cities. First is the one without those regions - smaller cities or the ones where public transport stops are not do dense to form a regions of this characteristic (an example is Poznan or Cologne). Second type have those hubs in city center, most often in the neighborhood of a train station (an example can be Gdansk, Gdynia and Zagreb in Figure \ref{fig:line-cities}). Final type is a city with multiple \textit{hub-like} regions which form bigger groups (see Berlin in Figure \ref{fig:berlin-clusters1} or Warsaw in Figure \ref{fig:imbalance-warsaw} as well as Madrid, Athens or Budapest which can be seen on a linked map). It represents cities with the highest availability of public transport or with a very dense public transport network in a city center.

Using the second level of an obtained typology, we were able to extract regions with similar role across the cities. In this part, we present those regions to show, that our method is efficient of matching regions extraction between cities. We focused on regions in a city center, which can serve different purposes in cities public transport network. 

\begin{figure}[h]
    \centering
    \includegraphics[width=0.2\textwidth]{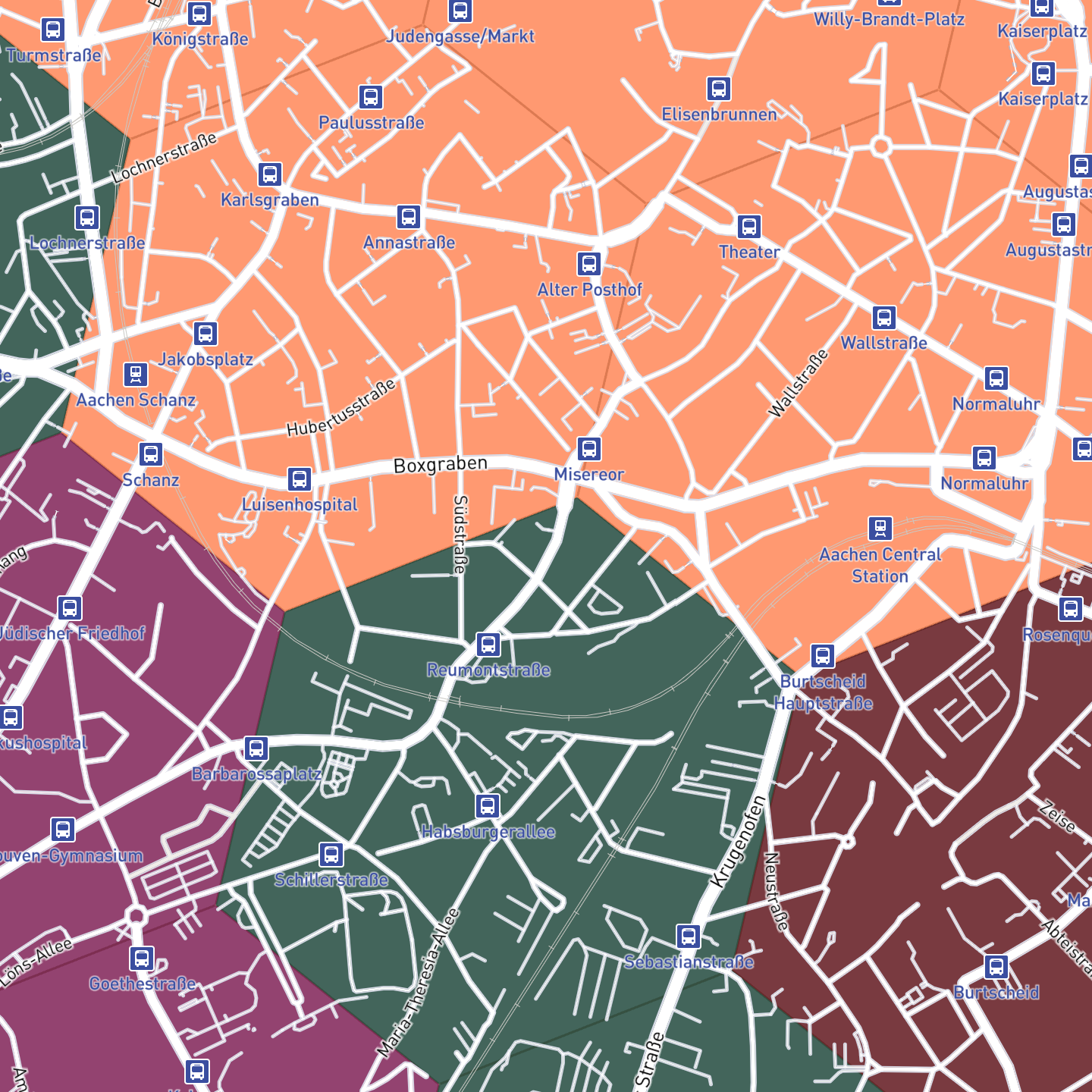}
    \hspace{0.2cm}
    \includegraphics[width=0.2\textwidth]{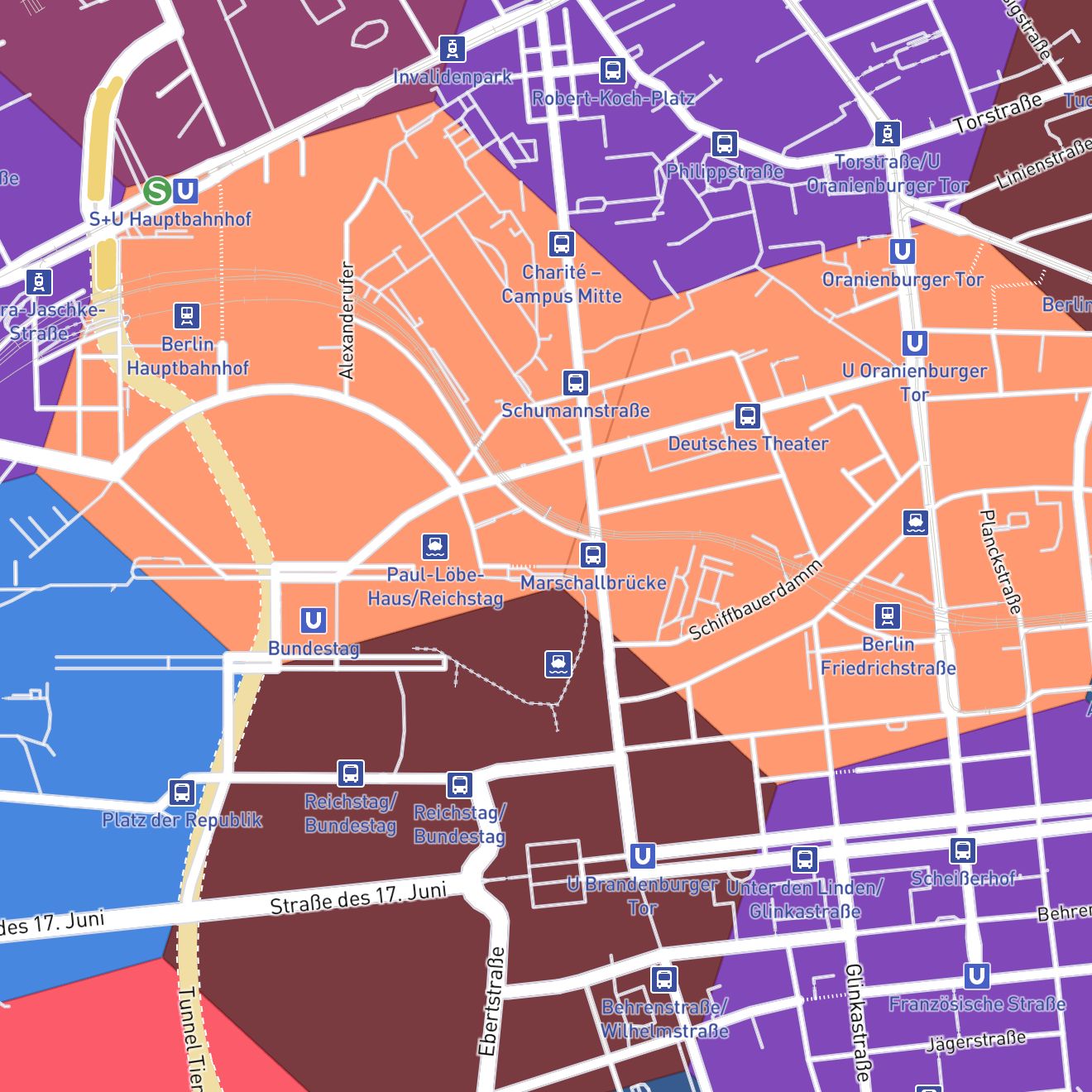}
    \includegraphics[width=0.2\textwidth]{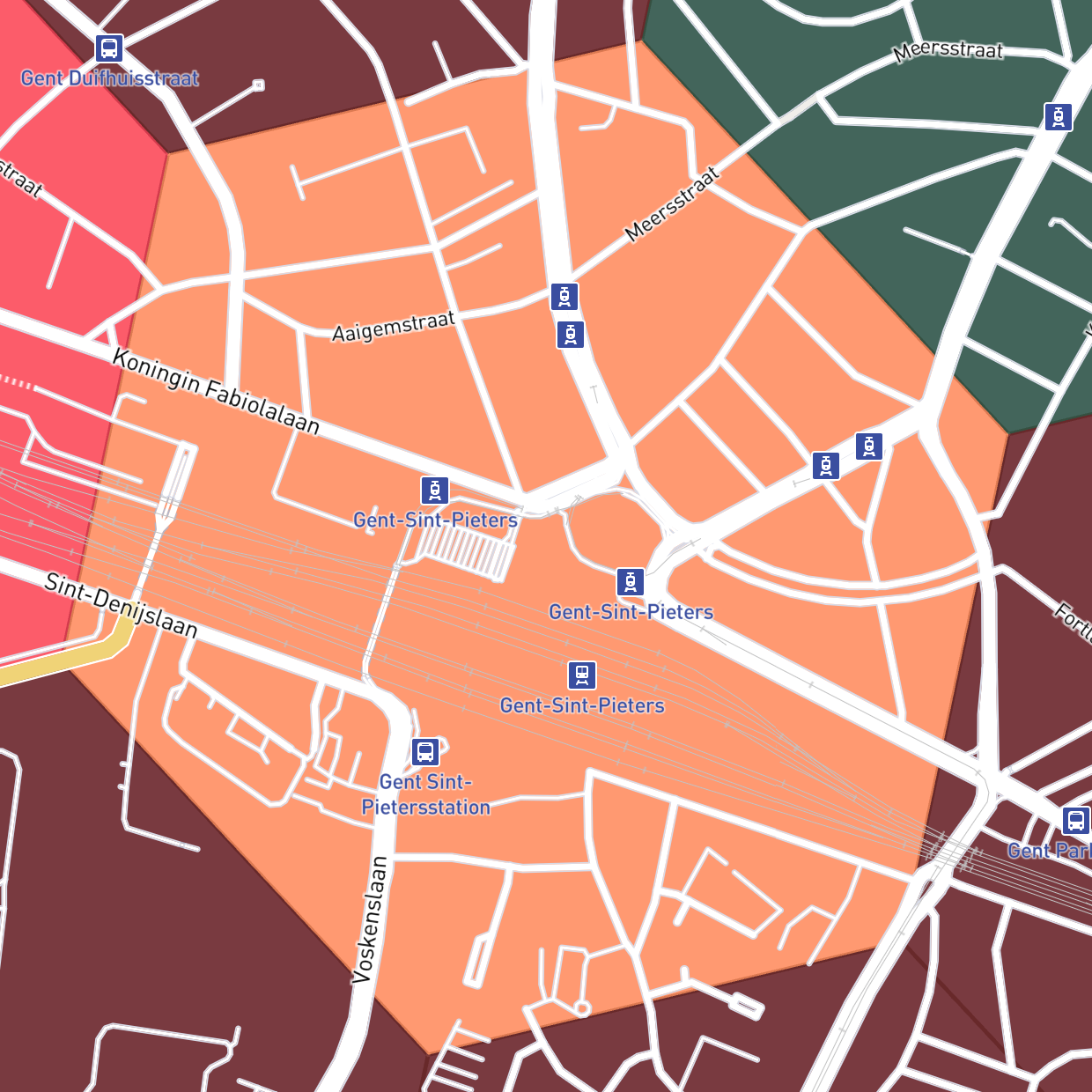}
    \hspace{0.2cm}
    \includegraphics[width=0.2\textwidth]{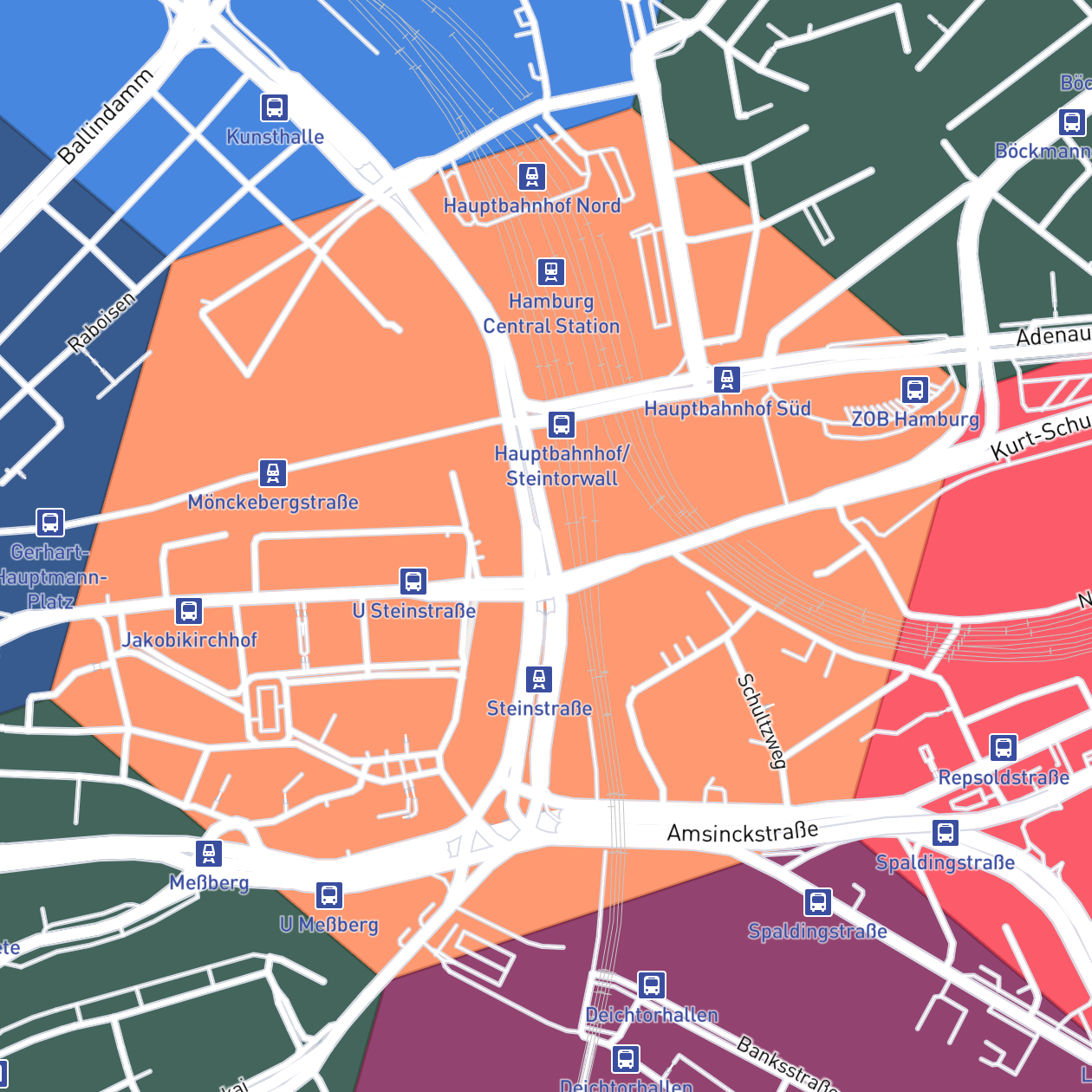}
    \includegraphics[width=0.2\textwidth]{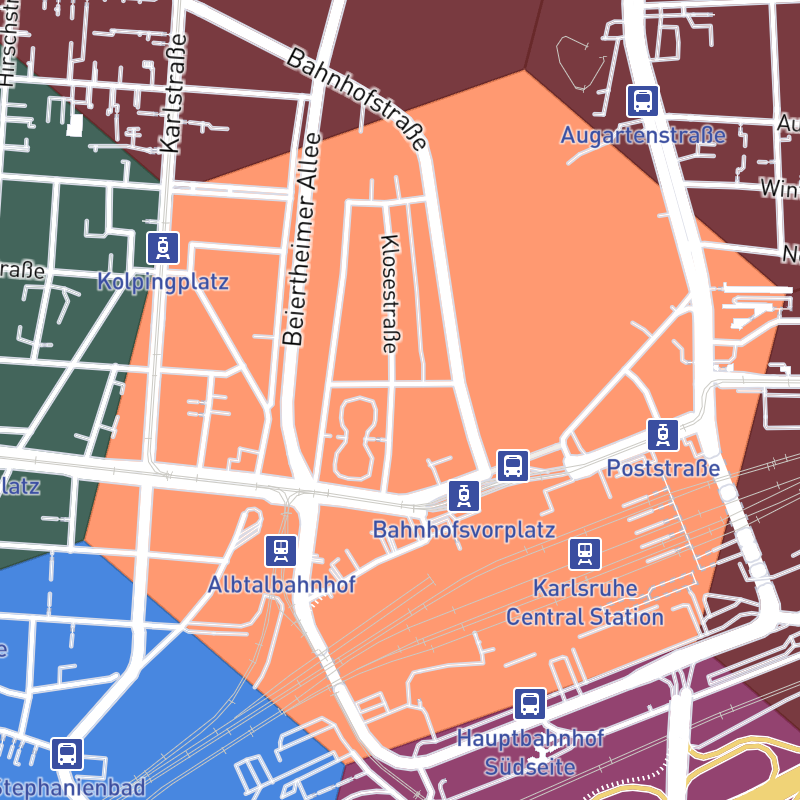}
    \hspace{0.2cm}
    \includegraphics[width=0.2\textwidth]{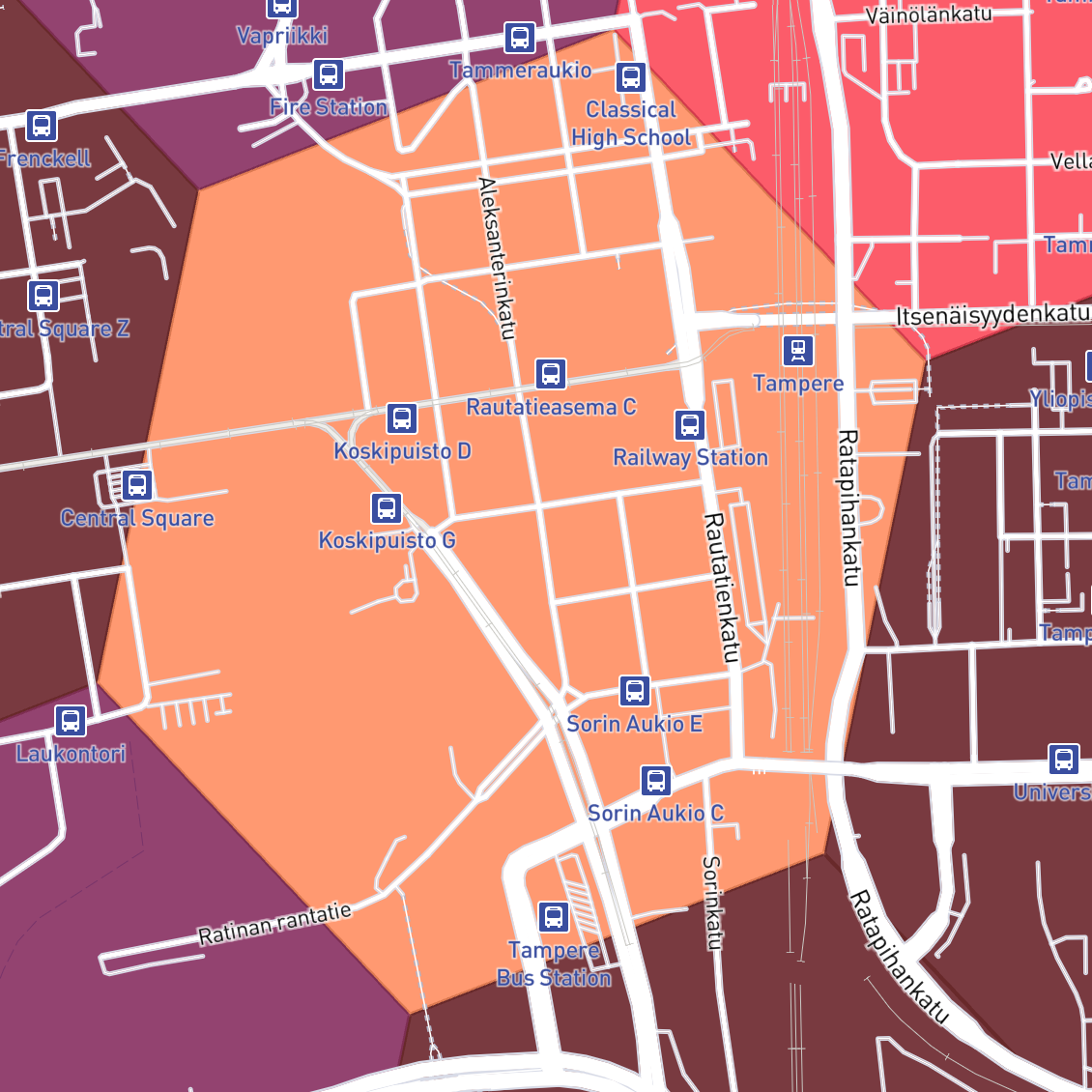}
    \caption{Regions with train stations inside are detected just based on public transport availability. From left to right: Aachen, Berlin, Gent, Hamburg, Karlsruhe and Tampere.}
    \label{fig:train-regions}
\end{figure}

First of the identified region types is a region with railway station in it. It was marked in this work as \textcolor{cl2-8}{cluster \textbf{8}} and in Figure \ref{fig:train-regions} we present examples from Aachen, Berlin, Gent, Hamburg, Karlsruhe and Tampere, where in each of those cities this cluster is assigned to the region with a train station inside. High availability of public transport in those regions is a combination of a railway connections and public transport stops in a close proximity. 

\begin{figure}[ht]
    \centering
    \includegraphics[width=0.2\textwidth]{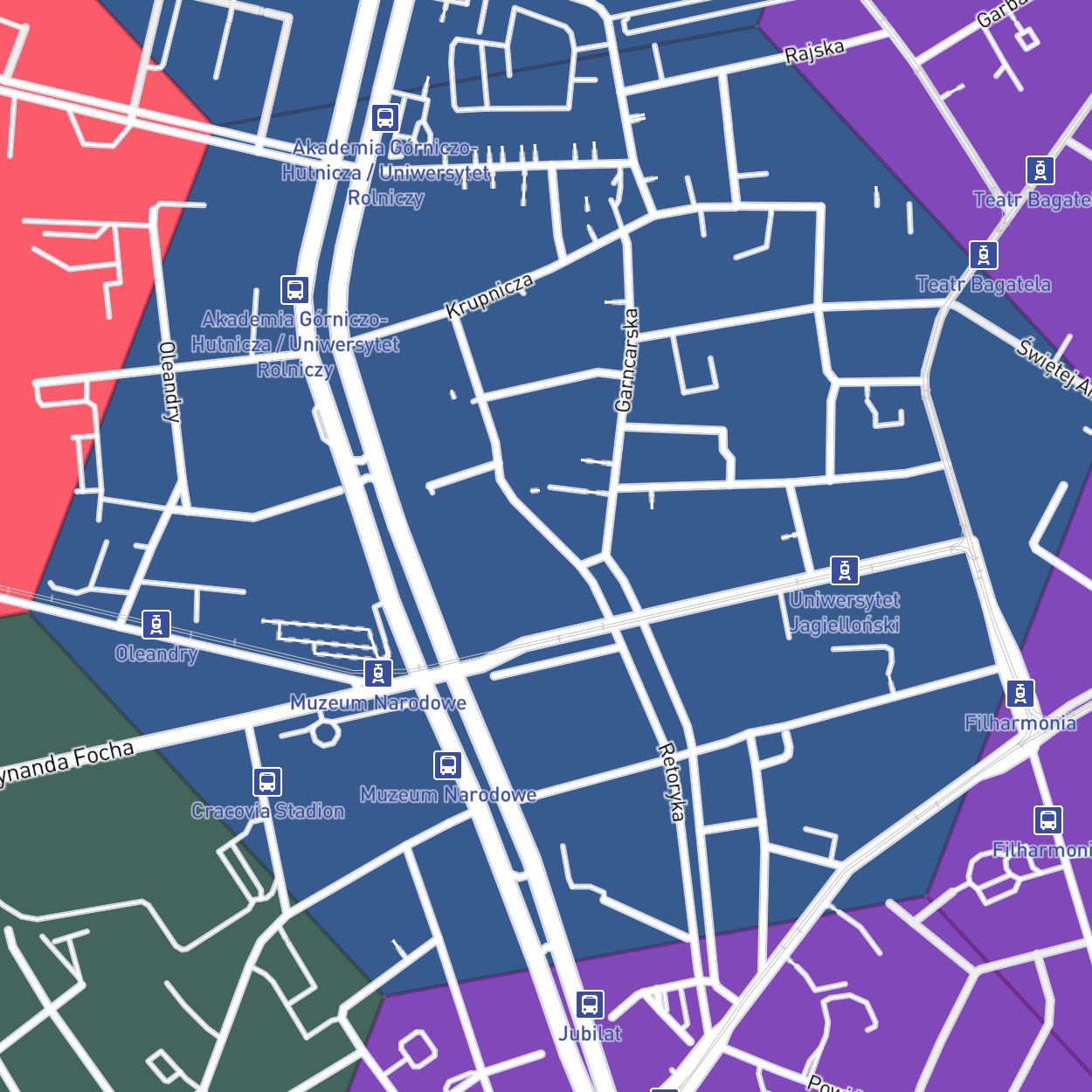}
    \hspace{0.2cm}
    \includegraphics[width=0.2\textwidth]{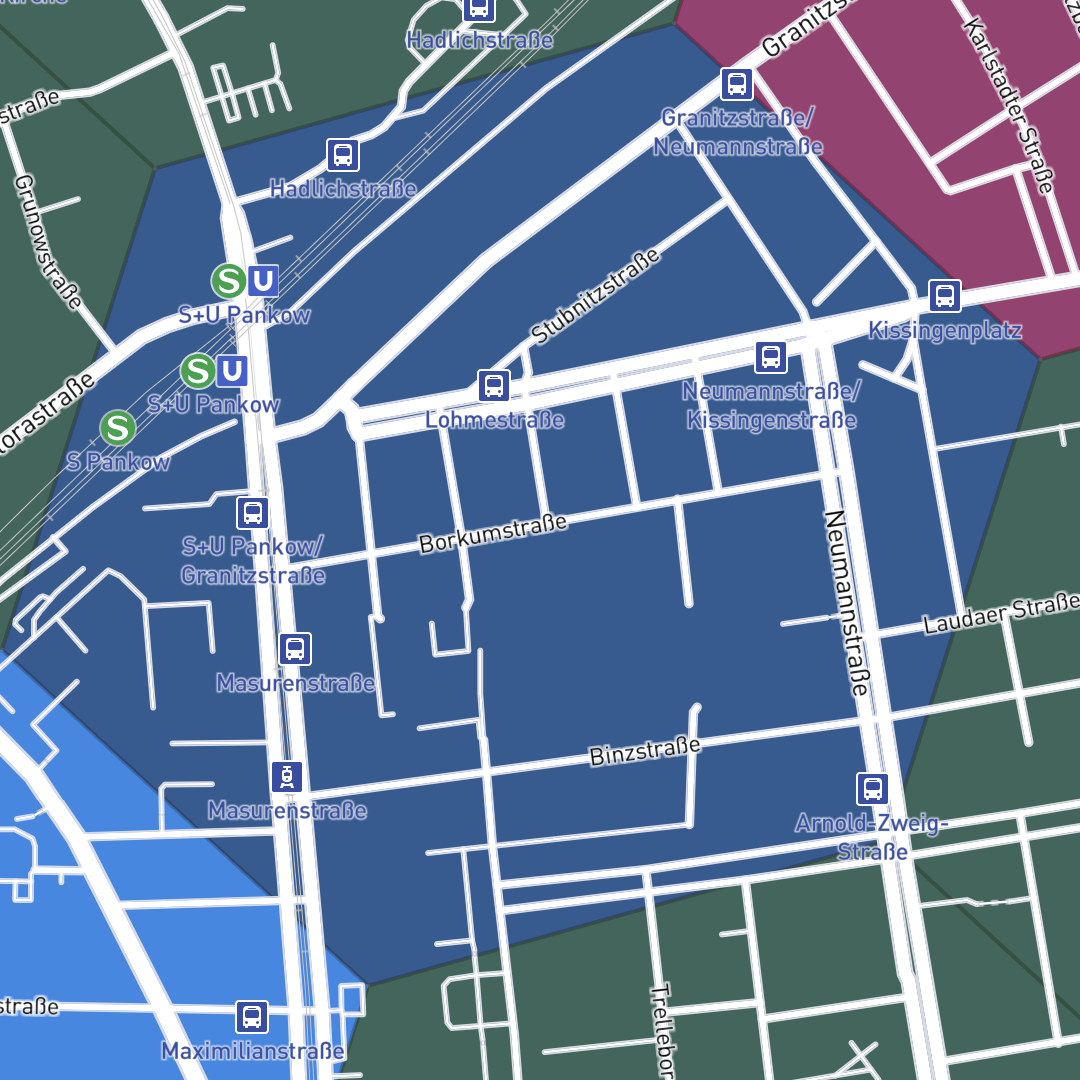}
    \includegraphics[width=0.2\textwidth]{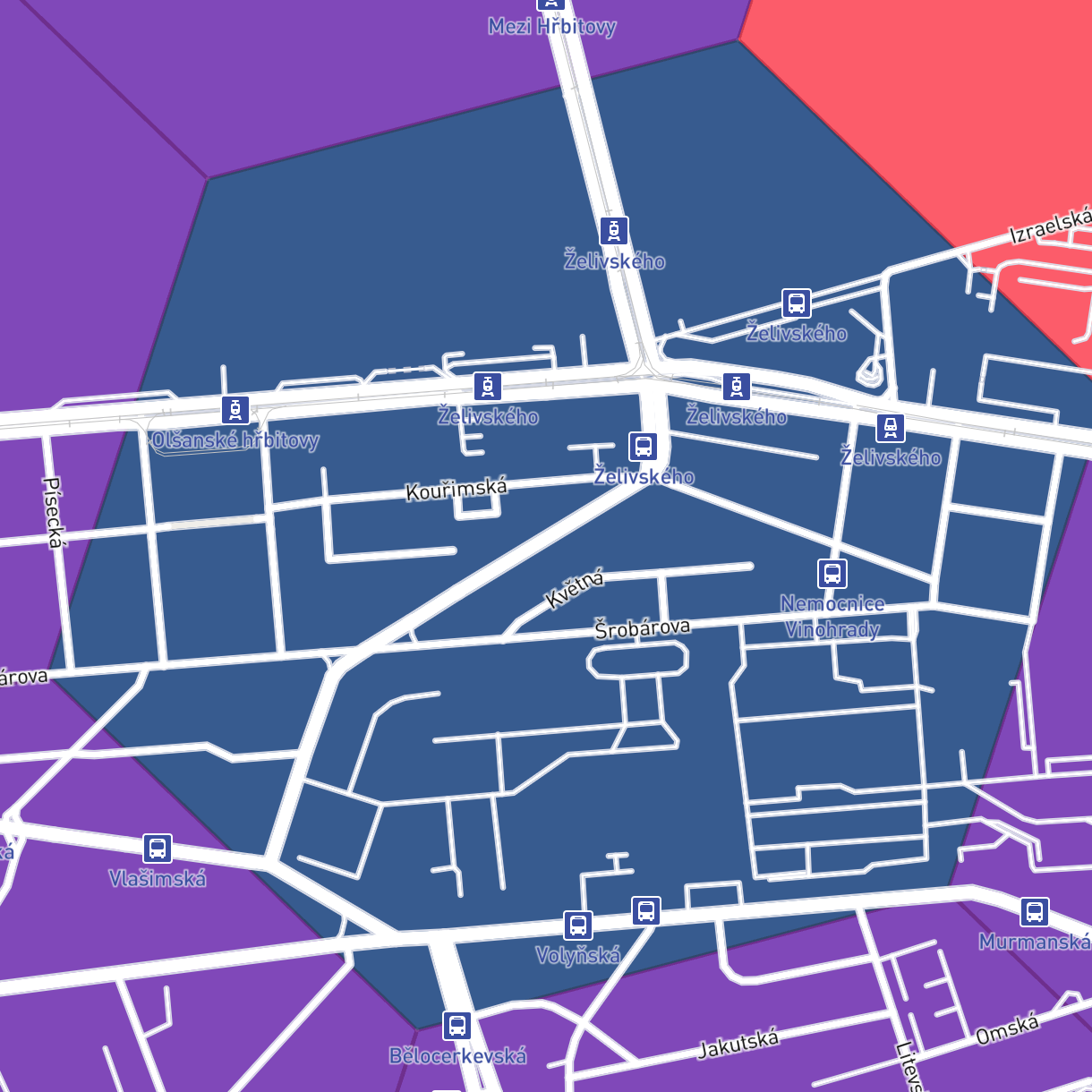}
    \hspace{0.2cm}
    \includegraphics[width=0.2\textwidth]{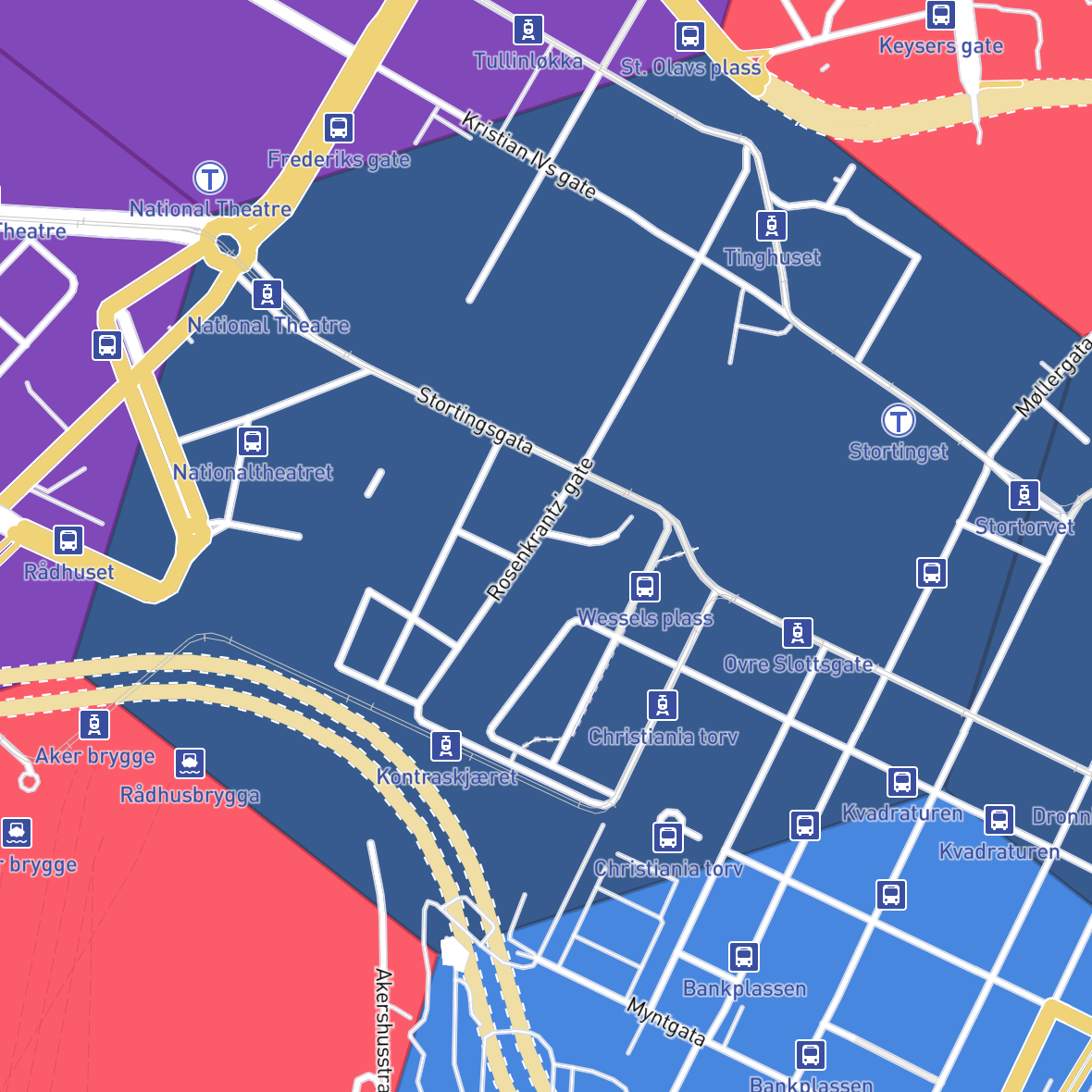}
    \caption{Regions from a center of cluster with high public transport availability and variety: Cracow, Berlin, Prague and Oslo.}
    \label{fig:super-hubs}
\end{figure}

Second type of region in a \textit{hub} category is represented by \textcolor{cl2-2}{cluster \textbf{2}} and combines regions with very high availability of public transport. In Figure \ref{fig:super-hubs} we presented 4 out of 5 regions from the center of \textcolor{cl2-2}{cluster \textbf{2}}. They land in Cracow, Berlin, Prague and Oslo and all contain regions with high number of stops of public transport available and serve the role of transportation hubs. 

Another type of region is a local hub associated with \textcolor{cl2-5}{cluster \textbf{5}}. Our analysis revealed, that those regions, when located further from a city center, serve a role of local transportation hubs/transfer points. Examples are shown in Figures \ref{fig:berlin-clusters2} (Berlin) and \ref{fig:local-hubs-wro} (Wroclaw). In Berlin they mostly match with U-Bahn or railway stations and in Wroclaw they cover major junctions where multiple public transport lines cross. 

\begin{figure}
    \centering
    \includegraphics[width=0.4\textwidth]{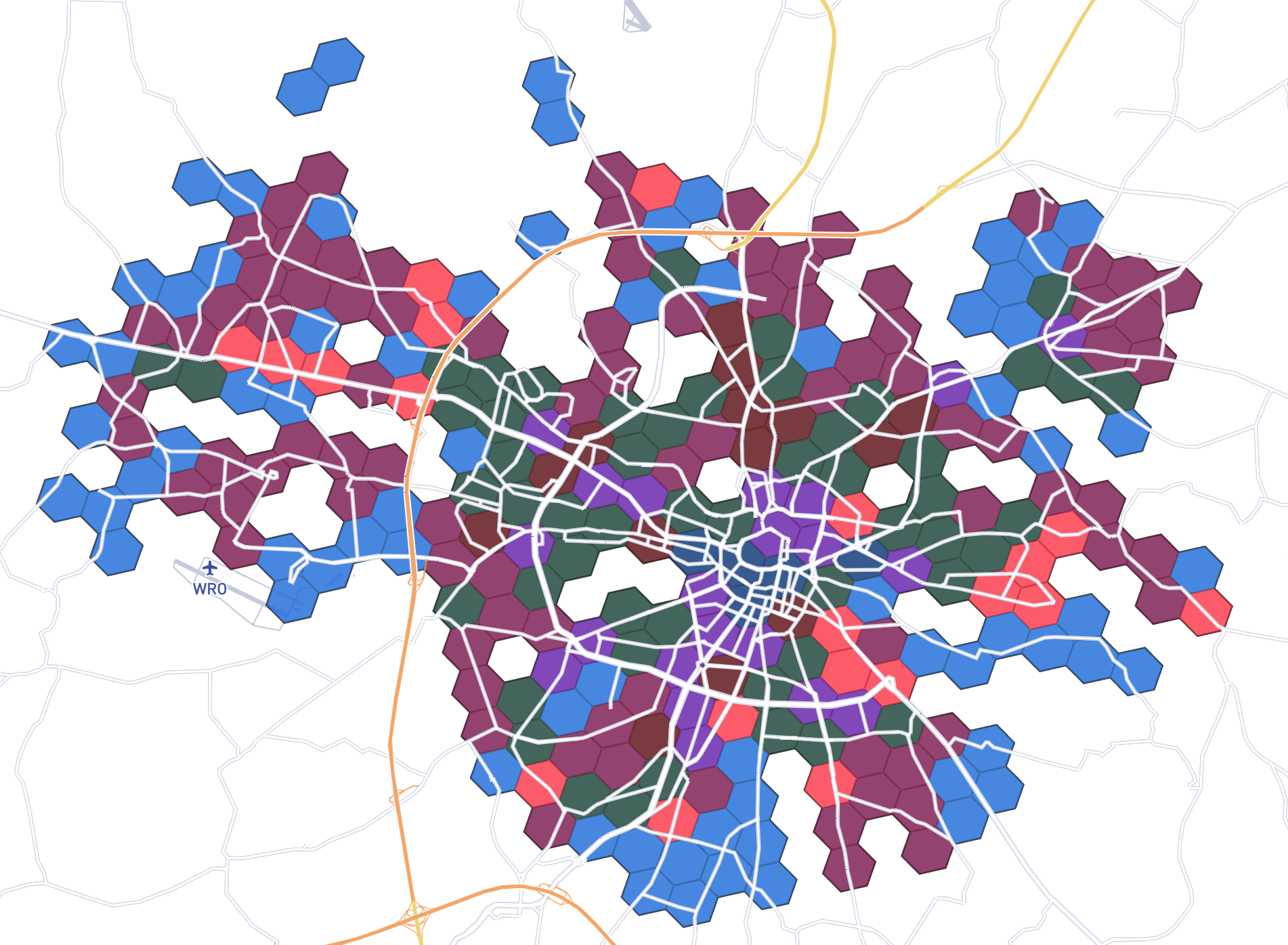}
    \caption{2nd level of typology visualized for Wroclaw, regions in \textcolor{cl2-5}{cluster \textbf{5}} with local hubs are clearly discernible}
    \label{fig:local-hubs-wro}
\end{figure}

The last example shows an analysis of a standing out city. Public transport network in Madrid is very dense and therefore it stands out collections most of the regions associated in \textcolor{cl2-6}{cluster \textbf{6}}. Almost whole city is classified to a type with high public transport availability. This is probably due to a very dense subway network in this city. This shows, that our proposed method is capable of not only identifying similar regions in different cities, but also finding the cities which stand out in terms of public transport availability. 

\begin{figure}
    \centering
    \includegraphics[width=0.4\textwidth]{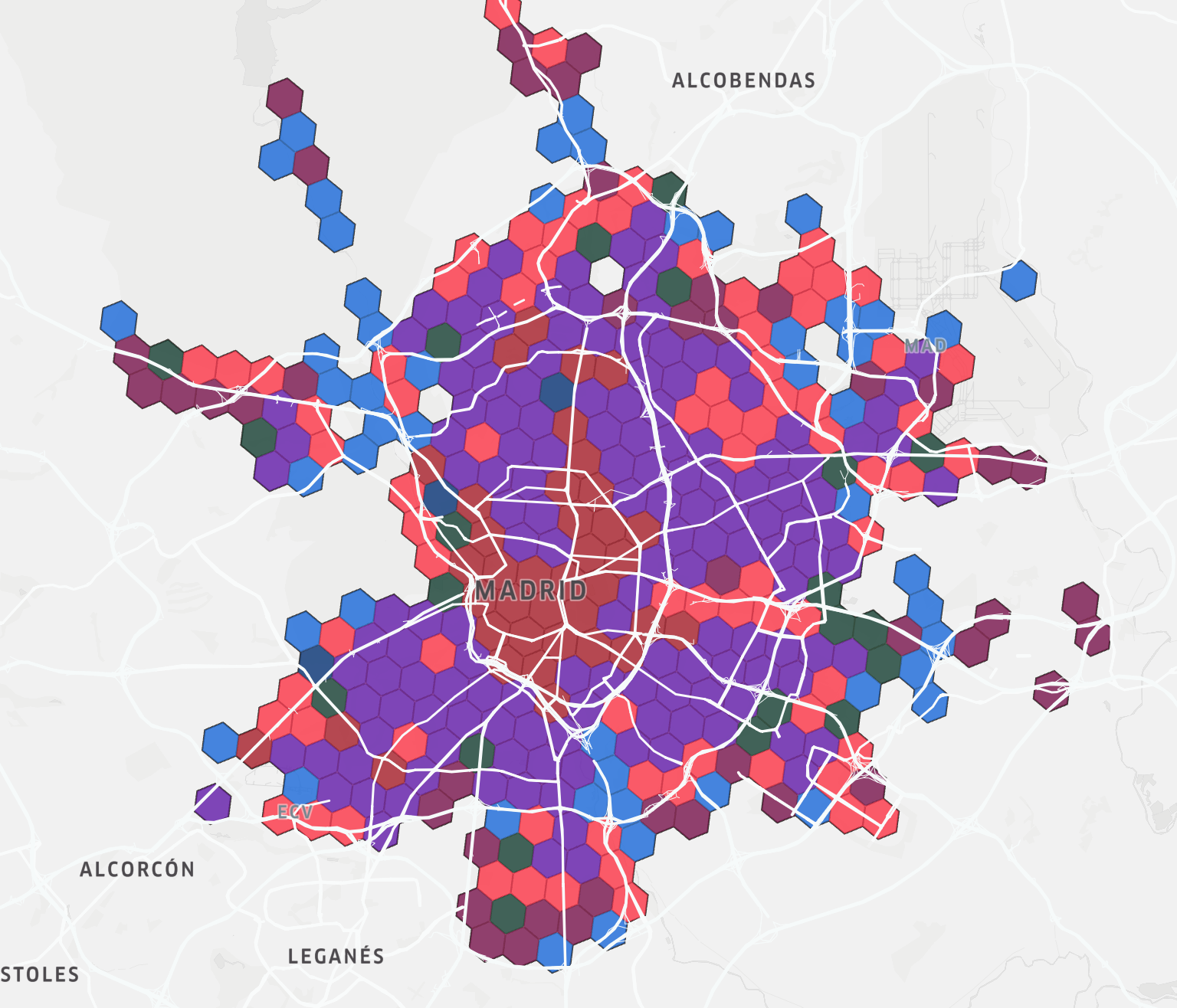}
    \caption{Madrid's dense public transport network in the city center - an (standing out) example of \textcolor{cl2-6}{cluster \textbf{6}} - regions with a lot of trips, but limited directions.}
    \label{fig:local-hubs-wro}
\end{figure}

\section{Conclusions and future work}

We proposed a solution to obtain vector representation of public transport offer in GTFS format. Using exploratory cluster analysis we show that it captures transport  characteristics and allows identifying similarities between regions. Thanks to using a wide selection of cities allowed for a good quality, our multi-level typology definition can be used for comparisons between cities and within them. Moreover, several unique functions of regions were identified in terms of their role in the public transport network. Finally we succesfully used obtained embeddings to search for regions with similar public transport offers across evaluated cities, which a novel contribution to the field. 

In the future investigating other forms of normalization can be an interesting direction. Usage of the local normalization approach, meaning that normalization is conducted separately for each city. This approach should allow for independence from the scale of available transportation in a city and allow for the identification of regions with similar functions despite different public transportation availability. At the cost of being able to compare cities among themselves, this approach should allow comparing the styles of public transport organizations in different cities.

\bibliographystyle{ACM-Reference-Format}
\bibliography{sample-sigconf}










\end{document}